\pdfoutput=1 

\documentclass[10pt,twocolumn,letterpaper]{article}

\usepackage{iccv}
\usepackage{times}
\usepackage{epsfig}
\usepackage{graphicx}
\usepackage{amsmath}
\usepackage{amssymb}
\usepackage{mathtools}
\usepackage{multicol}

\usepackage{threeparttablex}
\usepackage{longtable}
\usepackage[dvipsnames]{xcolor}

\usepackage{scalerel}
\usepackage{stmaryrd}
\usepackage{multirow}
\usepackage{subcaption}
\usepackage{float}
\usepackage{wrapfig}
\usepackage{makecell}
\usepackage{booktabs} 
\usepackage[font=small]{caption} 
\usepackage{sidecap}
\sidecaptionvpos{figure}{c}

\usepackage{subcaption}
\usepackage{graphicx}
\usepackage{lipsum}
\usepackage[normalem]{ulem}

\usepackage{listings}
\usepackage{xcolor}

\definecolor{codegreen}{rgb}{0,0.6,0}
\definecolor{codegray}{rgb}{0.5,0.5,0.5}
\definecolor{codepurple}{rgb}{0.58,0,0.82}
\definecolor{backcolour}{rgb}{0.95,0.95,0.92}

\lstdefinestyle{mystyle}{
    backgroundcolor=\color{backcolour},
    commentstyle=\color{codegreen},
    keywordstyle=\color{magenta},
    numberstyle=\tiny\color{codegray},
    stringstyle=\color{codepurple},
    basicstyle=\ttfamily\footnotesize,
    breakatwhitespace=false,
    breaklines=true,
    captionpos=b,
    keepspaces=false,
    numbers=left,
    numbersep=5pt,
    showspaces=false,
    showstringspaces=false,
    showtabs=false,
    tabsize=2,
    basicstyle=\ttfamily\scriptsize
}

\usepackage{hyperref}

\definecolor{mycolor}{HTML}{F7F8E0}
\definecolor{darkcyan}{RGB}{0,113,194}
\definecolor{forestgreen}{RGB}{34,139,34}

\hypersetup{colorlinks,breaklinks=true,
	citecolor=darkcyan
}

\iccvfinalcopy 

\def\iccvPaperID{9385} 

\ificcvfinal\fi

\begin{document}

\title{DocFormer: End-to-End Transformer for Document Understanding}

\author{Srikar Appalaraju \\
AWS AI \\
{\tt\small srikara@amazon.com}
\and
Bhavan Jasani\\
AWS AI\\
{\tt\small bjasani@amazon.com}
\and
Bhargava Urala Kota \\
AWS AI \\
{\tt\small bharkota@amazon.com}
\and
Yusheng Xie \\
AWS AI \\
{\tt\small yushx@amazon.com}
\and
R. Manmatha \\
AWS AI \\
{\tt\small manmatha@amazon.com}
}

\maketitle
\ificcvfinal\thispagestyle{empty}\fi

\newcommand{\xrightarrowdbl}[2][]{%
	\leftarrow\mathrel{\mkern-14mu}\xrightarrow[#1]{#2}
}
\newcommand{\papertitle}{DocFormer }
\renewcommand\theadset{\def\arraystretch{.85}}

\begin{abstract}
We present \papertitle- a multi-modal transformer based architecture for the task of Visual Document Understanding (VDU). VDU is a challenging problem which aims to understand documents in their varied formats (forms, receipts etc.) and layouts. In addition, \papertitle is pre-trained in an unsupervised fashion using carefully designed tasks which encourage multi-modal interaction. \papertitle uses text, vision and spatial features and combines them using a novel multi-modal self-attention layer. \papertitle also shares learned spatial embeddings across modalities which makes it easy for the model to correlate text to visual tokens and vice versa.  \papertitle is evaluated on 4 different datasets each with strong baselines. \papertitle achieves state-of-the-art results on all of them, sometimes beating models 4x its size (in no. of parameters). 
\end{abstract}

\vspace{-2.5em}
\section{Introduction}
The task of Visual Document Understanding (VDU) aims at understanding digital documents either born as PDF's or as images. VDU focuses on varied document related tasks like entity grouping, sequence labeling, document classification. While modern OCR engines \cite{litman2020scatter} have become good at predicting text from documents, VDU often requires understanding both the structure and layout of documents. The use of text or even text and spatial features alone is not sufficient for this purpose. For the best results, one needs to exploit the text, spatial features and the image. One way to exploit all these features is using transformer models  \cite{brown2020language,devlin2018bert,vaswani2017attention}. Transformers have recently been used for VDU  \cite{bros2020hong,xu2020layoutlm,xu2020layoutlmv2}. These models differ in how the unsupervised pre-training is done, the way self-attention is modified for the VDU domain or how they fuse modalities (text and/or image and spatial). There have been text only \cite{devlin2018bert}, text plus spatial features only \cite{bros2020hong,xu2020layoutlm} approaches for VDU. However, the holy-grail is to fuse all three modalities (text, visual and spatial features). This is desirable since there is some information in text that visual features miss out (language semantics), and there is some information in visual features that text misses out (text font and visual layout for example). 

\begin{figure}[t]
	\begin{center}
		\includegraphics[width=1.0\linewidth,width=230pt]{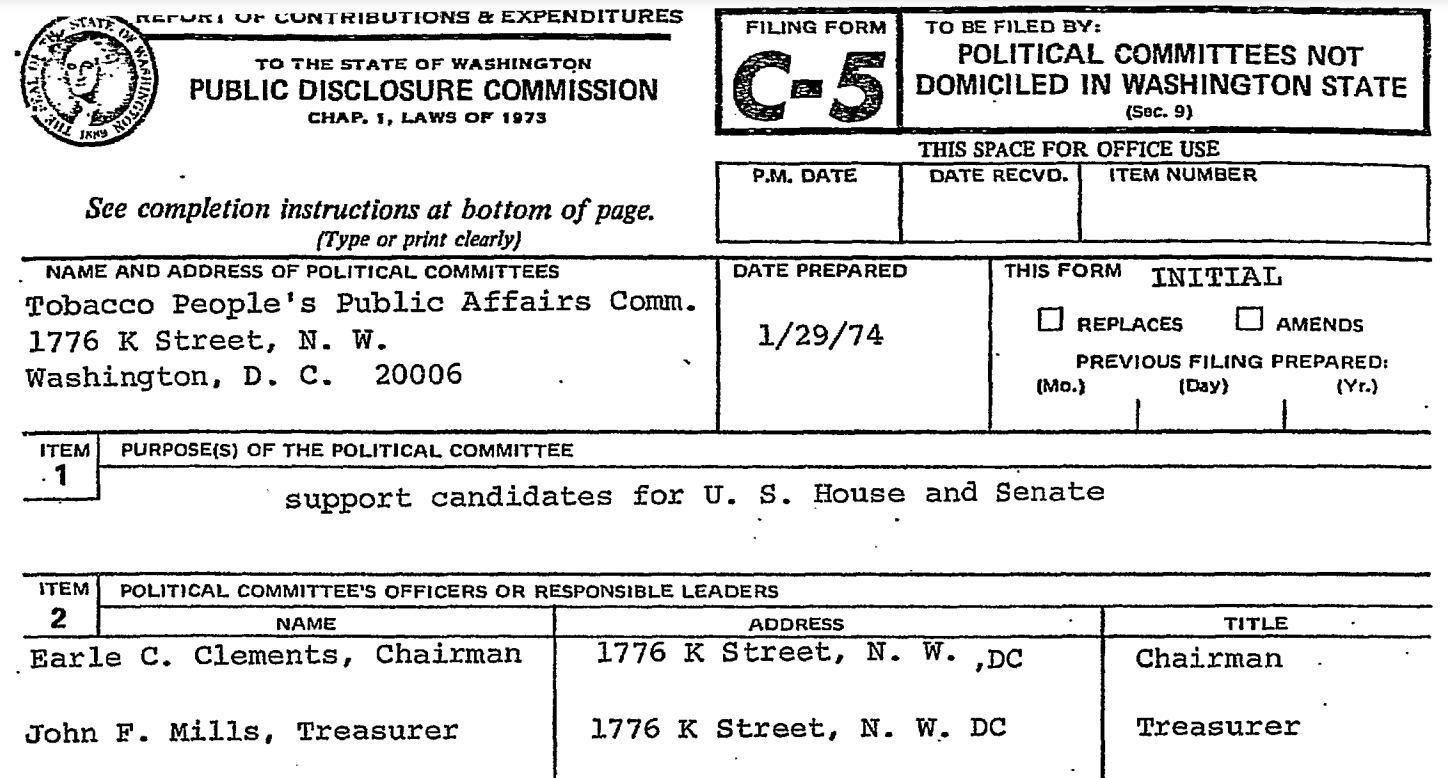}
	\end{center}
	\vspace{-3ex}
	\caption{\textbf{Snippet of a Document}: Various VDU tasks on this document may include labeling each text token into fixed classes or grouping tokens into a semantic class and finding relationships between tokens e.g. (``DATE PREPARED''  $\shortrightarrow$ Key and ``1/29/74'' $\shortrightarrow$ Value) or classifying the document into different categories. Note a document could have ``other'' text e.g. ``C-5'' which the model should ignore or classify as ``other'' depending on the task.  }
	\label{fig:splash_img}
	\vspace{-3ex}
\end{figure}

Multi-modal training in general is difficult since one has to map a piece of text to an arbitrary span of visual content. For example in Figure \ref{fig:splash_img}, ``ITEM 1'' needs to be mapped to the visual region.  Said a different way, text describes semantic high-level concept(s) e.g. the word ``person'' whereas visual features map to the pixels (of a person) in the image. It is not easy to enforce feature correlation across modalities from text $\xrightarrowdbl{\text{}}$ image. We term this issue as \textit{cross-modality feature correlation} and reference it later to show how \papertitle presents an approach to address this.

\papertitle follows the now common, pre-training and fine-tuning strategy. \papertitle incorporates a novel multi-modal self-attention with shared spatial embeddings in an encoder only transformer architecture. In addition, we propose three pre-training tasks of which two are novel unsupervised multi-modal tasks: \textit{learning-to-reconstruct} and \textit{multi-modal masked language modeling} task. Details are provided in Section \ref{section:approach}. To the best of our knowledge, this is the first approach for doing VDU which does not use bulky pre-trained object-detection networks for visual feature extraction. \papertitle instead uses plain ResNet50 \cite{he2016deep} features along with shared spatial (between text and image) embeddings which not only saves memory but also makes it easy for \papertitle to correlate text, visual features via spatial features. 
\papertitle is trained end-to-end with the visual branch trained from scratch. We now highlight the contributions of our paper:

\begin{itemize}
	\item A novel multi-modal attention layer capable of fusing text, vision and spatial features in a document. 
	\item Three unsupervised pre-training tasks which encourage multi-modal feature collaboration. Two of these are novel unsupervised multi-modal tasks: \textit{learning-to-reconstruct} task and a \textit{multi-modal masked language modeling} task. 
	\item \papertitle is end-to-end trainable and it does not rely on a pre-trained object detection network for visual features simplifying its architecture. On four varied downstream VDU tasks, \papertitle achieves state of the art results. On some tasks it out-performs large variants of other transformer almost 4x its size (in the number of parameters). In addition, \papertitle does not use custom OCR unlike some of the recent papers \cite{xu2020layoutlmv2,bros2020hong}.
\end{itemize}

\begin{figure}[t]
	\begin{center}
		\includegraphics[width=1.0\linewidth]{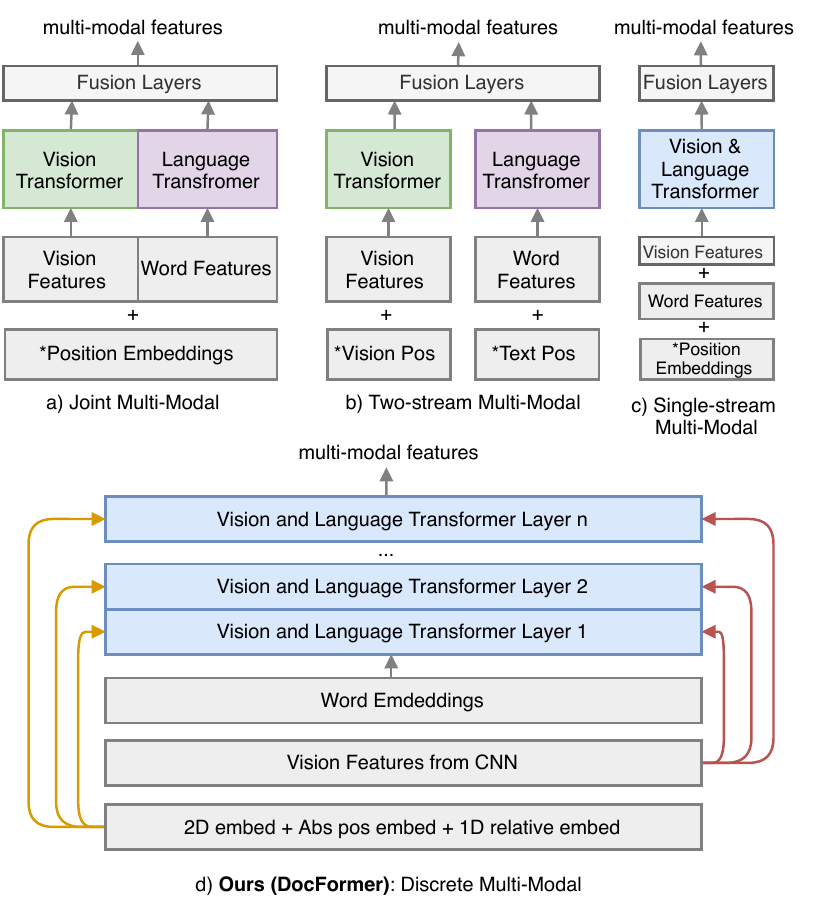}
	\end{center}
	\vspace{-3ex}
	\caption{Conceptual Comparisons of \textbf{Transformer Multi-Modal Encoder Architectures}: The mechanisms differ in how the modalities are combined. \textbf{Type A)} Joint Multi-Modal: like VL-BERT\cite{su2019vl}, LayoutLMv2\cite{xu2020layoutlmv2}, VisualBERT \cite{li2019visualbert}, MMBT\cite{kiela2019supervised}, UNITER \cite{chen2020uniter} \textbf{Type B)} Two-stream Multi-Modal: CLIP\cite{radford2021learning}, VilBERT\cite{lu2019vilbert}, \textbf{Type C)} Single-stream Multi-Modal, \textbf{Type D)} Ours: Discrete Multi-modal. e.g. \papertitle. Note: in each transformer layer, each input modality is self-attended separately. Best viewed in color.}
	\label{fig:encoder_methodologies}
	\vspace{-2.5ex}
\end{figure}



\section{Background}

Document understanding methods in the literature have used various combinations of image, spatial and text features in order to  understand and extract information from structurally rich documents such as forms~\cite{hammami2015one,zhou2016irmp,davis2019deep}, tables~\cite{schreiber2017deepdesrt,zhong2019publaynet,herzig2020tapas}, receipts~\cite{huang2019icdar2019,hong2020bros} and invoices~\cite{lohani2018invoice,riba2019table,majumder2020representation}. Finding the optimal way to combine these multi-modal features is an active area of research.

Grid based methods~\cite{katti2018chargrid,denk2019bertgrid} were proposed for invoice images where text pixels are encoded using character or word vector representations and classified into field types such as Invoice Number, Date, Vendor Name and Address etc. using a convolutional neural network. 

BERT~\cite{devlin2018bert} is a transformer-encoder~\cite{vaswani2017attention} based neural network that has been shown to work well on language understanding tasks. LayoutLM~\cite{xu2020layoutlm} modified the BERT architecture by adding 2D spatial coordinate embeddings along with 1D position and text token embeddings. They also added visual features for each word token, obtained using a Faster-RCNN and its bounding box coordinates.

LayoutLM was pre-trained on 11 million unlabeled pages and was then finetuned on several document understanding tasks - form processing, classification and receipt processing. This idea of pre-training on large datasets and then finetuning on several related downstream tasks is also seen in general vision and language understanding work \cite{su2019vl,lu2019vilbert,kiela2019supervised,li2019visualbert} etc. Figure \ref{fig:encoder_methodologies} shows a comparison of multi-modal transformer encoder architectures.

Recently, LayoutLMv2~\cite{xu2020layoutlmv2} improved over LayoutLM by changing the way visual features are input to the model - treating them as separate tokens as opposed to adding visual features to the corresponding text tokens. Further, additional pre-training tasks were explored to make use of unlabeled document data.


BROS~\cite{hong2020bros} also uses a BERT based encoder, with a graph-based classifier based on SPADE~\cite{hwang2020spatial}, which is used to predict entity relations between text tokens in a document. They also use 2D spatial embeddings added along with text tokens and evaluate their network on forms, receipts document images. Multi-modal transformer encoder-decoder architectures based on T5~\cite{raffel2019exploring} have been proposed recently. Tanaka et al. propose Layout-T5~\cite{tanaka2021visualmrc} for a question answering task on a database of web article document images whereas Powalski et al. propose TILT~\cite{powalski2021going} combining convolutional features with the T5 architecture to perform various downstream document understanding tasks.



\section{Approach}
\label{section:approach}

\textbf{Conceptual Overview}: We first present a conceptual overview of architectures used in Transformer Encoder Multi-Modal training, illustrated in Figure \ref{fig:encoder_methodologies}. \textbf{(a) Joint Multi-Modal}: VL-BERT~\cite{su2019vl}, LayoutLMv2~\cite{xu2020layoutlmv2}, VisualBERT~\cite{li2019visualbert}, MMBT~\cite{kiela2019supervised}: In this type of architecture, vision and text are concatenated into one long sequence which makes transformers self-attention hard due to the \textit{cross-modality feature correlation} referenced in the introduction. \textbf{(b) Two-Stream Multi-Modal} CLIP~\cite{radford2021learning}, VilBERT~\cite{lu2019vilbert}: It is a plus that each modality is a separate branch which allows one to use an arbitrary model for each branch. However, text and image interact only at the end which is not ideal. It might be better to do early fusion. \textbf{(c) Single-stream Multi-Modal}:  treats vision features also as tokens (just like language) and adds them with other features. Combining visual features with language tokens this way (simple addition) is unnatural as vision and language features are different types of data. (d) \textbf{Discrete Multi-Modal}: In this paper, \papertitle unties visual, text and spatial features. i.e. spatial and visual features are passed as residual connections to each transformer layer. We do this because spatial and visual dependencies might differ across layers. In each transformer layer, visual and language features separately undergo self-attention with shared spatial features. In order to pre-train \papertitle we use a subset of 5 million pages from the IIT-CDIP document collection \cite{lewis2006building} for pre-training. In order to do multi-modal VDU, we first extract OCR, which gives us text and corresponding word-level bounding boxes for each document. We next describe the model-architecture, followed by the pre-training tasks.





\subsection{Model Architecture}
\papertitle is an encoder-only transformer architecture. It also has a CNN backbone for visual feature extraction. All components are trained end-to-end. \papertitle enforces deep multi-modal interaction in transformer layers using novel multi-modal self-attention. We describe how three modality features (visual, language and spatial) are prepared before feeding them into transformer layers.

\textbf{Visual features}: 
Let $v \in \mathbb{R}^{3 \times h \times w}$ be the image of a document, which we feed through a ResNet50 convolutional neural network $f_{cnn}(\theta, v)$. We extract lower-resolution visual embedding at layer 4 i.e. $v_{l_{4}} \in \mathbb{R}^{c \times h_l \times w_l }$. Typical values at this stage are $c=2048$ and $h_l=\frac{h}{32}, w_l=\frac{w}{32}$ ($c$ = number of channels and $h_l$ and $w_l$ are the height and width of the features). The transformer encoder expects a flattened sequence as input of $d$ dimension. So we first apply a $1\times1$ convolution to reduce the channels $c$ to $d$. We then flatten the ResNet features to ($d$, $h_l \times w_l$) and use a linear transformation layer to further convert it to ($d$, $N$) where $d=768, N=512$. Therefore, we represent the visual embedding as $\overline{V} =  linear( conv_{1\times1} ( f_{cnn}(\theta, v) ) )$.




\textbf{Language features}: 
Let $t$ be the text extracted via OCR from a document image. In order to generate language embeddings, we first tokenize text $t$ using a word-piece tokenizer \cite{wu2016google} to get $t_{tok}$, this is then fed through a trainable embedding layer $W_{t}$. $t_{tok}$ looks like $[CLS], t_{tok_1}, t_{tok_2}, \dots, t_{tok_n}$ where $n=511$. If the number of tokens in a page is $> 511$, we ignore the rest. For a document with fewer than 511 tokens, we pad the sequence with a special $[PAD]$ token and we ignore the $[PAD]$ tokens during self-attention computation. We ensure that the text embedding, $\overline{T} =  W_t( t_{tok} )$, is of the same shape as the visual embedding $\overline{V}$. Following prior art \cite{xu2020layoutlmv2}, we initialize $W_{t}$ with LayoutLMv1 \cite{xu2020layoutlm} pre-trained weights.



\textbf{Spatial Features}: For each word $k$ in the text, we also get bounding box coordinates $b_k=(x_1,y_1, x_2,y_2, x_3,y_3, x_4,y_4)$. 2D spatial coordinates $b_{k}$ provide additional context to the model about the location of a word in relation to the entire document. This helps the model make better sense of the content. For each word, we encode the top-left and bottom-right coordinates using separate layers $W^{x}$ and $W^{y}$ for $x$ and $y$-coordinates respectively. We also encode more spatial features: bounding box height $h$, width $w$, the Euclidean distance from each corner of a bounding box to the corresponding corner  in the bounding box to its right and the distance between centroids of the bounding boxes, e.g. $A_{rel} = \{A_{num}^{k+1} - A_{num}^{k}\}; A \in (x,y); num \in (1,2,3,4,c)$, where $c$ is the center of the bounding box. Since transformer layers are permutation-invariant, we also use absolute 1D positional encodings $P^{abs}$. We create separate spatial embeddings for visual $\overline{V_{s}}$ and language $\overline{T_{s}}$ features since spatial dependency could be modality specific. Final spatial embeddings are obtained by summing up all intermediate embeddings. All spatial embeddings are trainable.


\vspace{-1.5em}
\begin{multline}
	\overline{V_{s}} = W_{v}^{x}(x_1,x_3,w,A_{rel}^x) + \\ W_{v}^{y}(y_1,y_3,h,A_{rel}^y) + P_{v}^{abs}
	\label{eqn:spatial_featuresl}
\end{multline}

\vspace{-2.9em}
\begin{multline}
	\overline{T_{s}} = W_{t}^{x}(x_1,x_3,w,A_{rel}^x) + \\ W_{t}^{y}(y_1,y_3,h,A_{rel}^y) + P_{t}^{abs}
	\label{eqn:lang_spatial_featuresl}
\end{multline}


\textbf{Multi-Modal Self-Attention Layer}: We now describe in detail our novel multi-modal self-attention layer. Consider a transformer encoder $f_{enc}(\eta, \overline{V}, \overline{V_{s}}, \overline{T}, \overline{T_{s}})$, where $\eta$ are trainable parameters of the transformer, $\overline{V}$, $\overline{V_{s}}$, $\overline{T}$ and $\overline{T_{s}}$ are visual, visual-spatial, language and language-spatial features respectively, and are obtained as described previously. Transformer $f_{enc}$ outputs a multi-modal feature representation $\overline{M}$ of the same shape $d=768, N=512$ as each of the input features. 

Self-attention, i.e., scaled dot-product attention as introduced in \cite{vaswani2017attention}, for a single head is defined as querying a dictionary with key-value pairs. i.e. in a transformer layer $l$ and $i^{th}$ input token in a feature length of $L$.



\vspace{-0.7em}
\begin{equation}
	\overline{M}_{i}^{l}=\sum_{j=1}^{L} \frac{\exp \left(\alpha_{i j}\right)}{\sum_{j^{\prime}=1}^{n} \exp \left(\alpha_{i j^{\prime}}\right)}\left(x_{j}^{l} W^{V, l}\right)
	\label{eqn:general_transformers}
\end{equation}

where $\alpha_{i j}$ is defined as self-attention which is computed as (attention in layer $l$ between tokens $x_i$ and $x_j$).

\vspace{-0.9em}
\begin{equation}
	\alpha_{i j}=\frac{1}{\sqrt{d}}\left(x_{i}^{l} W^{Q, l}\right)\left(x_{j}^{l} W^{K, l}\right)^{T}
	\label{eqn:self_attention}
\end{equation}

Here, $d$ is the dimension of the hidden representation, $W^{Q,l}, W^{K,l} \in \mathbb{R}^{d \times d_{K}}$, and $W^{V} \in \mathbb{R}^{d \times d_{V}}$ are learned parameter matrices which are not shared among layers or attention heads. Without loss of generality, we remove the dependency on layer l and get a simplified view of Eq. \ref{eqn:self_attention} as: 
\begin{equation}
	\alpha_{i j}=\left(x_{i} W^{Q}\right)\cdot\left(x_{j} W^{K}\right)^{T}
	\label{eqn:simple_self_attention}
\end{equation}

We modify this attention formulation for the multi-modal VDU task. \papertitle tries to infuse the following inductive bias into self-attention formulation: \textit{for most VDU tasks, local features are more important than global ones}. We modify Eq. \ref{eqn:simple_self_attention}, to add relative features. Specifically, the attention distribution for visual features is: 

\vspace{-1.5em}
\begin{multline}
	\alpha_{i j}^{v}= \underbrace{(x_{i}^{v} W^{Q}_{v})(x_{j}^{v} W^{K}_{v})^{T}}_{\text{key-query attn.}}  + \underbrace{(x_{i}^{v} W^{Q}_{v} a_{ij})}_{\text{query 1D relative attn.}} + \\ \underbrace{(x_{j}^{v} W^{K}_{v} a_{ij})}_{\text{key 1D relative attn.}} + \underbrace{(\overline{V_{s}}W_s^{Q})(\overline{V_{s}}W_s^{K})}_{\text{visual spatial attn.}}
	\label{eqn:visual_self_attention}
\end{multline}

Here, $x^{v}$ denotes visual features, $W^{K}_{v}, W^{Q}_{v}$ denote learnable matrices for key, query visual embeddings respectively. $W^{K}_{s}, W^{Q}_{s}$ denote learnable matrices for key, query spatial embeddings respectively. $a_{ij}$ is 1D relative position embedding between tokens $i,j$ i.e. $a_{ij} = W_{j-i}^{rel}$ where $W^{rel}$ learns how token $i$ attends to $j$. We clip the relative attention so \papertitle gives more importance to local features. We get a similar equation for language attention $\alpha_{i j}^{t}$: 

\vspace{-1.5em}
\begin{multline}
	\alpha_{i j}^{t}= (x_{i} W^{Q}_{t})(x_{j} W^{K}_{t})^{T} + (x_{i} W^{Q}_{t} a_{ij})~ + \\ (x_{j} W^{K}_{t} a_{ij}) + (\overline{T_{s}}W_s^{Q})(\overline{T_{s}}W_s^{K})
	\label{eqn:lang_self_attention}
\end{multline}

Here, $x$ is the output of the previous encoder layer, or word embedding layer if $l=1$. An important aspect of Eq. \ref{eqn:visual_self_attention} and Eq. \ref{eqn:lang_self_attention} is that we share spatial weights in each layer. i.e. the spatial attention weights ($W_s^{Q}, W_s^{K}$) are shared across vision and language. This helps the model correlate features across modalities.


Using the visual self-attention computed using Eq. \ref{eqn:visual_self_attention} in Eq. \ref{eqn:general_transformers}, gets us spatially aware, self-attended visual features $\hat{V_l}$. Similarly using Eq. \ref{eqn:lang_self_attention} in Eq. \ref{eqn:general_transformers}, gets us language features $\hat{T_l}$. The multi-modal feature output is given by $\overline{M}_l = \hat{V_l} + \hat{T_l}$. It should be noted that for layers $l > 1$, features $x$ in Eq. \ref{eqn:lang_self_attention} are multi-modal because we combine visual and language features at the output of layer $l-1$. The final $\overline{M}_{12}$ is consumed by downstream linear layers. 




\textbf{Why do multi-modal attention this way?} 
We untie the visual and spatial information and pass them to each layer of transformer. We posit that making visual and spatial information accessible across layers acts as an information residual connection \cite{he2016identity,veit2016residual} and is beneficial for generating superior multi-modal feature representation hence better addressing the issue of \textit{cross-modality feature correlation}. This is verified in our experiments (Section \ref{section:experiments}), where we show that \papertitle obtains state-of-the-art performance even when compared to models having four times the number of the parameters in some cases.
Further, sharing spatial weights across modalities in each layer gives \papertitle an opportunity to learn cross-modal spatial interactions while also reducing the number of parameters. In Sec. \ref{section:experiments}, we show that \papertitle is the smallest amongst its class of models, yet it is able to show superior performance. Code in supple.


\textbf{Run-time Complexity}:  The  run-time complexity of \papertitle is of the same order as that of the original self-attention model \cite{vaswani2017attention} (for details see supplemental material)

\subsection{Pre-training}
\begin{figure}[t]
	\begin{center}
		\includegraphics[width=1.0\linewidth]{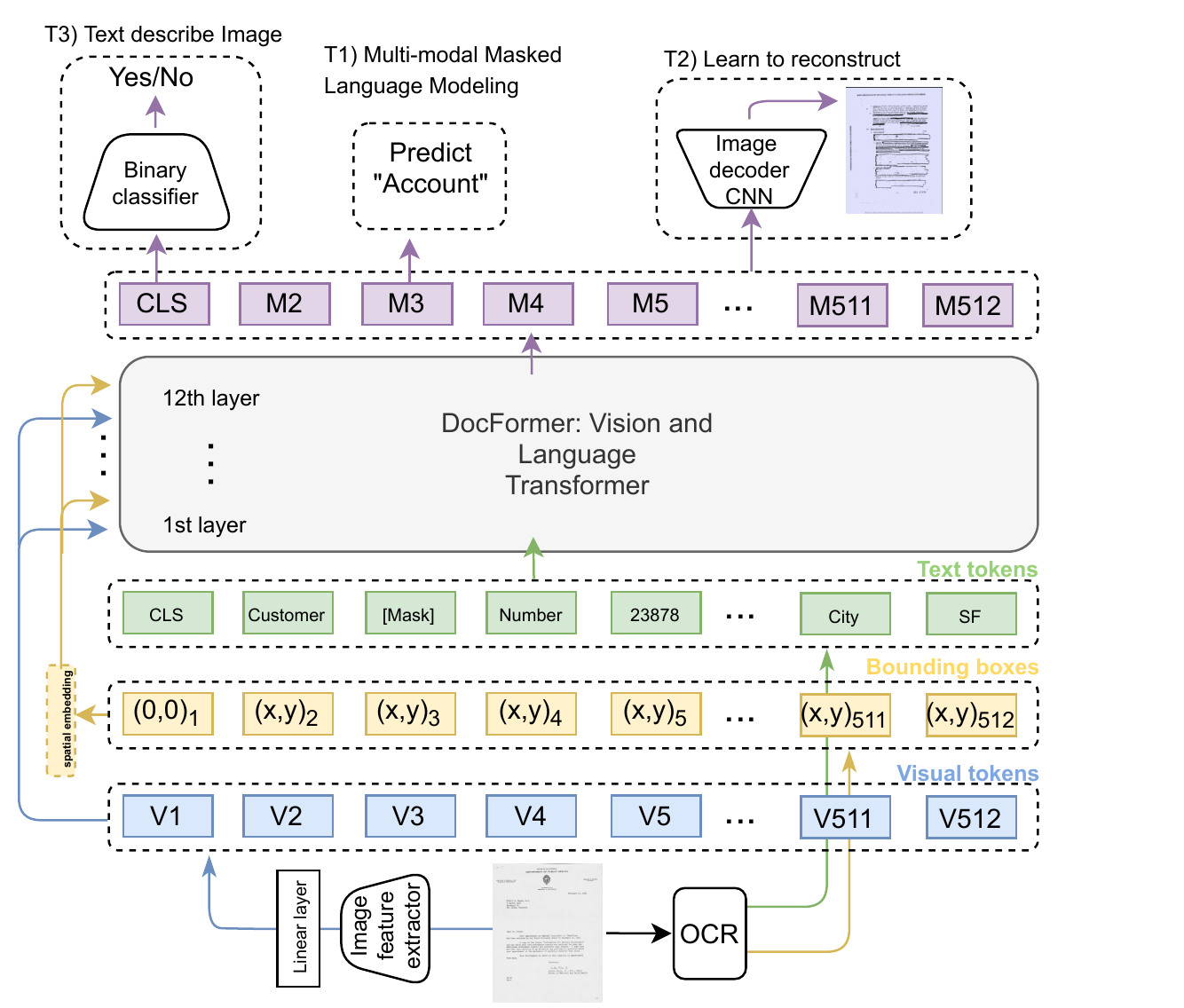}
	\end{center}
	\vspace{-3ex}
	\caption{\papertitle \textbf{pre-training methodology}. High level overview. Note: First bounding box token corresponding to [CLS], is meant for entire page coordinates.}
	\label{fig:docubert_pretrain}
	\vspace{-3ex}
\end{figure}

The ability to design new and effective unsupervised pre-training strategies is still an open problem. Our pre-training process involves passing the document image, its extracted OCR text, and its corresponding spatial features. All pre-training tasks were designed such that the network needs the collaboration of both visual and language features, thereby truly learning a superior representation than training with only one of the modalities. See Figure \ref{fig:docubert_pretrain} for a high-level overview of the pre-training tasks.
	


\textbf{Multi-Modal Masked Language Modeling (MM-MLM)}: This is a modification of the original masked language modeling (MLM) pre-text task  introduced in BERT \cite{devlin2018bert}, and may be thought of as a text de-noising task i.e. for a text sequence $t$, a corrupted sequence is generated $\widetilde{t}$. The transformer encoder predicts $\hat{t}$ and is trained with an objective to reconstruct entire sequence. In our case, we use a multi-modal feature embedding $\overline{M}$ for reconstruction of the text sequence. In prior art \cite{xu2020layoutlmv2, xu2020layoutlm}, for a masked text token, the corresponding visual region was also masked to prevent ``cheating''. Instead, we intentionally do not mask visual regions corresponding to $[MASK]$ text. This is to encourage visual features to supplement text features and thus minimize the text reconstruction loss. The masking percentage is the same as originally proposed \cite{devlin2018bert}. Cross-entropy loss is used for this task ($L_{MM-MLM}$).
	
\textbf{Learn To Reconstruct (LTR)}: In this novel pre-text task, we do the image version of the MM-MLM task, i.e. we do an image reconstruction task. The multi-modal feature predicted by \papertitle is passed through a shallow decoder to reconstruct the image (the same dimension as the input image). In this case this task is similar to an auto-encoder image reconstruction but with multi-modal features. The intuition is that in the presence of both image and text features, the image reconstruction would need the collaboration of both modalities. We employ a smooth-L1 loss between the reconstructed image and original input image ($L_{LTR}$). 
	
\textbf{Text Describes Image (TDI)}: In this task, we try to teach the network if a given piece of text describes a document image. For this, we pool the multi-modal features using a linear layer to predict a binary answer. This task differs from the above two tasks in that this task infuses the global pooled features into the network (as opposed to MM-MLM and LTR focusing purely on local features). In a batch, 80\% of the time the correct text and image are paired, for the remaining 20\% the wrong image is paired with the text. A binary cross-entropy loss ($L_{TDI}$) is used for this task. Since the 20\% negative pair scenario interferes with the LTR task (for a  text $\xrightarrowdbl{\text{}}$ image pair mismatch the pair reconstruction loss would be high), the LTR loss is ignored for cases where there is a mismatch.

The final pre-training loss $L_{pt} = \lambda L_{MM-MLM} + \beta L_{LTR} + \gamma L_{TDI}$. In practice $\lambda=5, \beta=1$ and $\gamma=5$. \papertitle is pre-trained for 5 epochs, then we remove all three task heads. We add one linear projection head and fine-tune all components of the model for all downstream tasks.

\section{Experiments}
\label{section:experiments}


For all experiments, we fine-tune on the training set and report numbers on the corresponding test/validation dataset. No dataset specific hyper-parameter tuning was done. We treat this as a plus and our reported numbers could be higher if dataset specific fine-tuning was done. For all downstream tasks, we use the official provided annotations unless otherwise stated. A common theme amongst these datasets is the relatively small amount of training data (most with $<$1000 samples). We posit that pre-training is much more helpful in such scenarios and helps measure the generalization capability of \papertitle. 

\textbf{Notations}: Tables \ref{table:funsd}, \ref{table:rvl_cdip}, \ref{table:cord}, \ref{table:nda}, use the following notation. T: Text features, S: spatial features. I: image features. \textbf{Bold} indicates SOTA. \underline{Underline} indicates second best. $\dagger$ denotes the use of Encoder + Decoder transformer layers. $*$ signifies approximate estimation.

\textbf{Implementation details}: We summarize details for pre-training and fine-tuning in Table 1 in supplemental. 
We emphasize the importance of warm-up steps and learning rate scale. 
We found that these settings have a non-trivial impact on pre-training result as well as downstream task performance. 
We used Pytorch \cite{paszke2019pytorch} and the Huggingface library \cite{wolf2019huggingface}.

\textbf{Models}: We employ the commonly used terminology for transformer encoder models - \textit{base} with 12 transformer layers (768 hidden state and 12 attention heads) and large with 24 transformer layers (1024 hidden state and 16 attention heads). We show that \papertitle-base gets SOTA for three of the 4 tasks beating even large models and for the 4th task is close to a large model. In addition to the multi-modal \papertitle, we also present a text and spatial \papertitle 
by pre-training \papertitle multi-modally but fine-tuning with only text and spatial features. We do this to show the flexibility of our model and show that during pre-training visual features were infused into \papertitle leading it to do better than pure text and spatial models.

\subsection{Sequence Labeling Task}
FUNSD \cite{jaume2019} dataset is a form understanding task. It contains 199 noisy documents (149 train, 50 test) which are scanned and annotated. We focus on the semantic entity-labeling task (i.e., group tokens which belong to the same class). 
We measure entity-level performance using F1 score shown in Table \ref{table:funsd}. \papertitle-base  achieves 83.34\% F1 score which is better than comparable models: LayoutLMv2-base \textcolor{forestgreen}{(+0.58)}, BROS \textcolor{forestgreen}{(+2.13)}, LayoutLMv1-base \textcolor{forestgreen}{(+4.07)}. Story repeats for \papertitle-large inspite of it trained only with 5M pages.

\begin{table}[htbp]
	\centering
	\scalebox{0.8}{
	\begin{tabular}{l|c|c|c|c}
		Model & \#param (M) & Precision & Recall & F1 \\ 
		\Xhline{2.5\arrayrulewidth}
		
		\multicolumn{5}{l}{\textit{methods based on only text / (text + spatial) features:}} \\ \midrule
		BERT-base \cite{devlin2018bert} & 109 & 54.69 & 61.71 & 60.26  \\
		RoBERTa-base \cite{liu2019roberta} & 125 & 63.49 & 69.75 & 66.48 \\
		UniLMv2-base  \cite{bao2020unilmv2} & 125 & 63.49 & 69.75 & 66.48  \\ 
		LayoutLMv1-base \cite{xu2020layoutlm} & 113 & 76.12 & 81.55 & 78.66  \\
		BROS-base \cite{bros2020hong} & 139 & 80.56 & 81.88 & 81.21  \\ 
		\midrule
		BERT-large \cite{devlin2018bert} & 340 & 61.13 & 70.85 & 65.63 \\
		RoBERTa-large \cite{liu2019roberta} & 355 & 67.80 & 73.91 & 70.72 \\
		UniLMv2-large  \cite{bao2020unilmv2} & 355 & 67.80 & 73.91 & 70.72  \\ 
		LayoutLMv1-large \cite{xu2020layoutlm} & 343 & 75.36 & 80.61 & 77.89  \\
		\midrule
		\multicolumn{5}{l}{\textit{methods based on image + text + spatial features:}} \\ \midrule
		LayoutLMv1-base  \cite{xu2020layoutlm} & 160 & 76.77 & 81.95 & 79.27  \\
		LayoutLMv2-base \cite{xu2020layoutlmv2}  & 200 & 80.29 & 85.39 & 82.76  \\ 
		LayoutLMv2-large \cite{xu2020layoutlmv2}  & 426 & 83.24 & 85.19 & \uline{84.20}  \\ 
		\midrule
		DocFormer-base (T+S) & 149 & 77.63 & 83.69 & 80.54 \\ 
		DocFormer-base (I+T+S) & 183 & 80.76 & 86.09 & 83.34 \\ 
		DocFormer-large (T+S) & 536 & 81.33 & 85.44 & 83.33 \\
		\textbf{DocFormer-large (I+T+S)} & 536 & 82.29 & 86.94 & \textbf{84.55} \\
		
	\end{tabular}
	}
	\caption{\textbf{FUNSD comparison}: \papertitle does better than  models its size and compares well with even larger models }
	\label{table:funsd}
	\vspace{-2.5ex}
\end{table}

\begin{SCfigure}[10]
  \centering
  \includegraphics[scale=0.55]{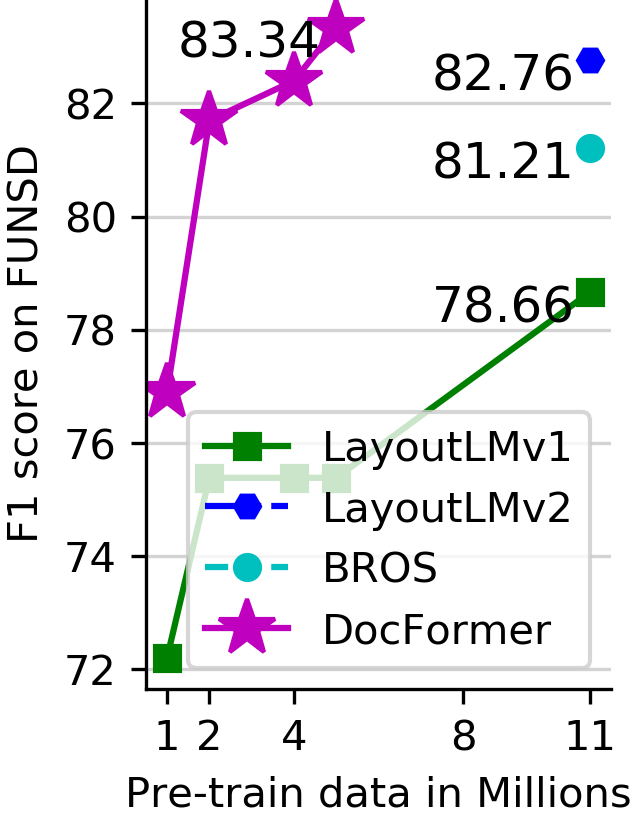}
  \caption{\textbf{Amount of Pre-training matters}: $x$-axis is the number of pre-training samples needed. $y$-axis is the F1-score on FUNSD task. \papertitle-base gets 83.34 after pre-training on only 5M pages and outperforms current SOTA LayoutLMv2-base's 82.76 which was pretrained on more than 2x more data.}
  \label{fig:docformer_funsd_epochs}
  \vspace{-3ex}
\end{SCfigure}

\textbf{FUNSD performance vs Pre-training samples}: We also measure the performance of \papertitle-base with increasing number of pre-training samples. As seen in Figure \ref{fig:docformer_funsd_epochs}, our base model achieves state-of-the-art performance of 83.34 F1-score in-spite of being pre-trained with only 5M documents. Previous SOTA needed more than 2x pre-training documents (11M) to achieve (82.76). Also \papertitle converges faster. 

\textbf{\papertitle performance without images}: Please note \papertitle-base T+S model which was pre-trained with (I+T+S) but was fine-tuned on FUNSD without Images gives F1 of 80.54 which is \textcolor{forestgreen}{+1.88\%} higher than a LayoutLMv1 (78.66\%) which was purely pre-trained and fine-tuned on T+S. We hypothesize that \papertitle was infused with visual features during pre-training and is better than text-only pre-trained models. 

\subsection{Document Classification Task}
For this task we use pooled features to predict a classification label for a document. The RVL-CDIP \cite{harley2015icdar} dataset consists of 400,000 grayscale images in 16 classes, with 25,000 images per class. Overall there are 320,000 training images, 40,000 validation images, and 40,000 test images. We report performance on test and eval metric is the overall classification accuracy. In line with prior art \cite{xu2020layoutlmv2,bros2020hong} text and layout information is extracted using Textract OCR. \papertitle-base achieves state-of-the-art performance of 96.17\%. \papertitle gives superior performance to all existing base and large transformer variants. Some models greater than 4x in number of parameters (TILT-large, 780M parameters gives 94.02\% (\textcolor{red}{-2.15}\% gap). 

\begin{table}[htbp]
	\centering
	\scalebox{0.8}{
		\begin{tabular}{l|c|c}
			Model & \#param (M) & Accuracy (\%) \\ 
			\Xhline{2.5\arrayrulewidth}
			
			\multicolumn{3}{l}{\textit{methods based on only images:}} \\ \midrule
			CNN ensemble \cite{harley2015icdar} & *60 & 89.80 \\
			VGG-16 \cite{afzal2017cutting} & 138 & 88.33 \\
			AlexNet \cite{tensmeyer2017analysis} & 61 & 90.94 \\
			GoogLeNet \cite{csurka2016right} & 13 & 90.70 \\ 
			Single Vision model \cite{das2018document} & *140 & 91.11 \\
			Ensemble \cite{das2018document} & - & 92.21 \\ 
			InceptionResNetV2 \cite{szegedy2017inception} & 56 & 92.63 \\
			LadderNet \cite{sarkhel2019deterministic} & - & 92.77 \\
			\midrule
			\multicolumn{3}{l}{\textit{methods based on text / (text + spatial) features:}} \\ 
			\midrule
			BERT-base \cite{devlin2018bert} & 110 & 89.81  \\
			UniLMv2-base  \cite{bao2020unilmv2} & 125 & 90.06 \\ 
			LayoutLMv1-base \cite{xu2020layoutlm} & 113 & 91.78	 \\
			
			BROS-base $\dagger$  \cite{bros2020hong} & 139 & 95.58  \\ 
			\midrule
			BERT-large \cite{devlin2018bert} & 340 & 89.92 \\
			UniLMv2-large  \cite{bao2020unilmv2} & 355 & 90.20 \\ 
			LayoutLMv1-large \cite{xu2020layoutlm} & 343 & 91.90  \\
			
			\midrule
			\multicolumn{3}{l}{\textit{methods based on image + text + spatial features:}} \\ 
			\midrule
			Single Modal \cite{dauphinee2019modular} & - & 93.03  \\
			Ensemble \cite{dauphinee2019modular} & - & 93.07 \\
			TILT-base $\dagger$ \cite{powalski2021going} & 230 & 93.50 \\
			LayoutLMv1-base \cite{xu2020layoutlm} & 160 & 94.42	 \\
			LayoutLMv2-base \cite{xu2020layoutlmv2} & 200 & 95.25  \\ 
			\midrule
			LayoutLMv1-large \cite{xu2020layoutlm} & 390 & 94.43	 \\
			TILT-large $\dagger$ \cite{powalski2021going}  & 780 & 94.02 \\
			LayoutLMv2-large \cite{xu2020layoutlmv2} & 426 & \underline{95.65}  \\ 
			\midrule
			\textbf{DocFormer-base }(I+T+S) & 183 & \textbf{96.17}  \\
			DocFormer-large (I+T+S) & 536 & 95.50
			
		\end{tabular}
	}
	\caption{\textbf{RVL-CDIP dataset} \cite{harley2015icdar} \textbf{comparison}: We report classification accuracy on the test set. \papertitle gets the highest classification accuracy and outperforms TILT-large by \textcolor{forestgreen}{+2.15} which is almost 4x its size.}
	\label{table:rvl_cdip}
\end{table}

\begin{table}[htbp]
	\scalebox{0.8}{
		\begin{tabular}{l|c|c|c|c}
			Model & \#param (M) & Precision & Recall & F1 \\ 
			\Xhline{2.5\arrayrulewidth}
			
			\multicolumn{5}{l}{\textit{methods based on only text / (text + spatial) features:}} \\ \midrule
			BERT-base \cite{devlin2018bert} & 109 & 88.33 & 91.07 & 89.68  \\
			UniLMv2-base  \cite{bao2020unilmv2} & 125 & 89.87 & 91.98 & 90.92  \\ 
			SPADE \cite{hwang2020spatial} & - & - & - & 91.50 \\
			LayoutLMv1-base \cite{xu2020layoutlm} & 113 & 94.37 & 95.08 & 94.72  \\
			BROS-base $\dagger$ \cite{bros2020hong}   & 139 & 95.58 & 95.14 & 95.36  \\ 
			\midrule
			BERT-large \cite{devlin2018bert} & 340 & 88.86 & 91.68 & 90.25 \\
			UniLMv2-large  \cite{bao2020unilmv2} & 355 & 91.23 & 92.89 & 92.05 \\ 
			LayoutLMv1-large \cite{xu2020layoutlm} & 343 & 94.32 & 95.54 & 94.93 \\
		
			\midrule
			\multicolumn{5}{l}{\textit{methods based on image + text + spatial features:}} \\ \midrule
			LayoutLMv2-base \cite{xu2020layoutlmv2}  & 200 & 94.53 & 95.39 & 94.95  \\ 
			TILT-base $\dagger$ \cite{powalski2021going} & 230 & - & - & 95.11 \\
			LayoutLMv2-large \cite{xu2020layoutlmv2} & 426 & 95.65 & 96.37 & 96.01 \\
			TILT-large $\dagger$ \cite{powalski2021going} & 780 & - & - & \uline{96.33} \\
			\midrule
			DocFormer-base (T+S) & 149 & 94.82 & 95.07 & 94.95\\ 
			DocFormer-base (I+T+S) & 183 & 96.52 & 96.14 & \uline{96.33} \\
			DocFormer-large (T+S) & 502 & 96.46 & 96.14 & 96.30\\ 
			\textbf{DocFormer-large} (I+T+S) & 536 & 97.25 & 96.74 & \textbf{96.99} \\
		\end{tabular}
	}
	\caption{\textbf{CORD dataset} \cite{park2019cord} \textbf{comparison}. We present entity-level Precision, Recall, F1 on test set.}
	\label{table:cord}
\end{table}

\vspace{-1em}
\subsection{Entity Extraction Task}
We report performance on two different entity extraction datasets: 

\textbf{CORD Dataset}~\cite{park2019cord}: consists of receipts. It defines 30 fields under 4 categories. The task is to label each word to the right field. The evaluation metric is entity-level F1. We use the provided OCR annotations and bounding boxes for fine-tuning (Table \ref{table:cord}). \papertitle-base achieves 96.33\% F1 on this dataset besting all prior *-base and virtually all *-large variants tying with TILT-large~\cite{powalski2021going} which has higher number of parameters. \papertitle-large achieves 96.99\% besting all other *-large variants achieving SOTA.

\textbf{Kleister-NDA}~\cite{gralinski2020kleister}: dataset consists of legal NDA documents. The task with Kleister-NDA data is to extract the values of four fixed labels. The approach needs to learn to ignore unrelated text. This dataset is challenging since it has some “decoy” text, for which no label should be given. Also, there might be more than one value given for a given label and all values need to be extracted. In line with prior-art we measure F1-score on validation data (since ground truth is not provided for test data). Also we extract OCR and apply heuristics to create train/validation ground-truth on OCR (Table \ref{table:nda}).

\begin{table}[htbp]
	\centering
	\scalebox{0.8}{
		\begin{tabular}{l|c|c}
			Model & \#param (M) & F1 \\ 
			\Xhline{2.5\arrayrulewidth}
			
			\multicolumn{3}{l}{\textit{methods based on only text / (text + spatial) features:}} \\ \midrule
			LAMBERT \cite{garncarek2020lambert} & - & 75.7 \\
			RoBERTa \cite{liu2019roberta} & 125 & 76.7 \\
			BERT-base \cite{devlin2018bert} & 110 & 77.9 \\
			UniLMv2-base  \cite{bao2020unilmv2} & 125 & 79.5 \\ 
			LayoutLMv1-base \cite{xu2020layoutlm} & 113 & 82.7 \\
			\midrule
			BERT-large \cite{devlin2018bert} & 340 & 79.1 \\
			UniLMv2-large  \cite{bao2020unilmv2} & 355 & 81.8 \\ 
			LayoutLMv1-large \cite{xu2020layoutlm} & 343 & 83.4 \\

			\midrule
			\multicolumn{3}{l}{\textit{methods based on image + text + spatial features:}} \\ \midrule
			LayoutLMv2-base \cite{xu2020layoutlmv2}  & 200 & 83.3  \\ 
			LayoutLMv2-large \cite{xu2020layoutlmv2} & 426 & \uline{85.2} \\
			\midrule
			\papertitle-base (T+S) & 149 & 82.1 \\ 
			\textbf{\papertitle-base }(I+T+S) & 183 & \textbf{85.8} \\
		\end{tabular}
	}
	\caption{\textbf{Kleister-NDA dataset} \cite{gralinski2020kleister} \textbf{comparison}: We present entity-level Precision, Recall, F1 on validation set. \papertitle gives the best performance, out-performing other *-large models trained with 2.5x the learning capacity.}
	\label{table:nda}
\end{table}

\subsection{More Experiments}
We conduct further analysis on the behavior of \papertitle pertaining to pre-training tasks, network structure and spatial embedding weight sharing.

\textbf{\textit{Shared or Independent Spatial embeddings?}} One of the benefits of our proposed \papertitle multi-modal self-attention architecture (Fig. \ref{fig:encoder_methodologies} and Eq. \ref{eqn:visual_self_attention},\ref{eqn:lang_self_attention}) is that sharing spatial embeddings across vision and language makes it easier for the model to learn feature-correlation across modalities. We see ablation on this aspect in Table \ref{table:w_wo_shared_spatial}. 
\begin{table}[H]
	\begin{center}
			\scalebox{0.75}{
		\begin{tabular}{l|c|c|c}
			Configuration & Num Params & FUNSD (F1) & CORD (F1) \\
			\Xhline{2.5\arrayrulewidth}
			w. shared spatial Eq. \ref{eqn:visual_self_attention},\ref{eqn:lang_self_attention} & 183 M & 76.9 & 93.36 \\
			w/o shared spatial & 198 M  & 75.58 \textcolor{red}{(-1.32)} & 92.51 \textcolor{red}{(-0.85)} \\
		\end{tabular}
	}
	\end{center}
	\vspace{-3ex}
	\caption{\textbf{Spatial Weight Sharing}: 
	In w/o shared spatial, vision and language get their own spatial weights $W_s$.}
	\label{table:w_wo_shared_spatial}
\end{table}

\textbf{\textit{Do our pre-training tasks help?}}  Pretraining is essential for  \textit{low-to-medium} data regimes (FUNSD and CORD). but even for downstream tasks with a lot of training samples (RVL-CDIP) it helps to improve performance and convergence (Table \ref{table:with_and_without_pretrain}). 

\begin{table}[H]
	\begin{center}
		\scalebox{0.75}{
			\begin{tabular}{l|c|c|c}
				\thead{Dataset} & \thead{Train\\ samples} &\thead{with pre-train \\ then 100 epochs (F1)} & \thead{w/o pre-train \\ 100 epochs (F1)} \\
				\Xhline{2.5\arrayrulewidth}
				FUNSD \cite{jaume2019} & 149 & 83.34 & 4.18 \\ 
				CORD \cite{park2019cord} & 800 & 96.33 & 0.54 \\ 
				RVL-CDIP \cite{harley2015icdar} & 320,000 & 96.17 & 93.95 \\
			\end{tabular}
		}
	\end{center}
	\vspace{-3ex}
	\caption{\textbf{Effect of Pre-training} 
	}
	\label{table:with_and_without_pretrain}
\end{table}



\begin{figure*}
	\centering
	
	
	
	

	\subcaptionbox{Ground Truth}[.288\linewidth][c]{%
		\includegraphics[width=\linewidth]{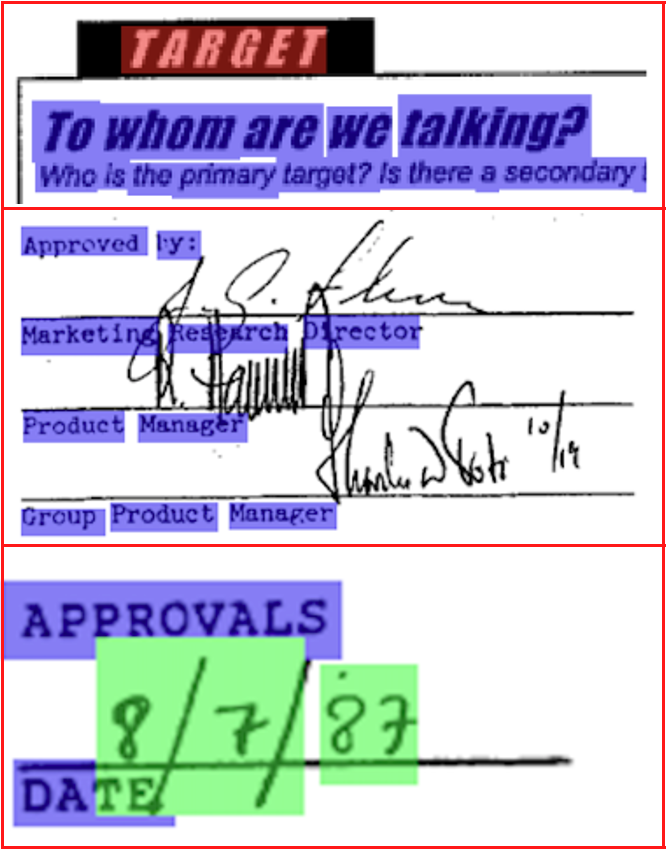}}\quad
	\subcaptionbox{Text + Spatial model \cite{xu2020layoutlm}}[.294\linewidth][c]{%
		\includegraphics[width=\linewidth]{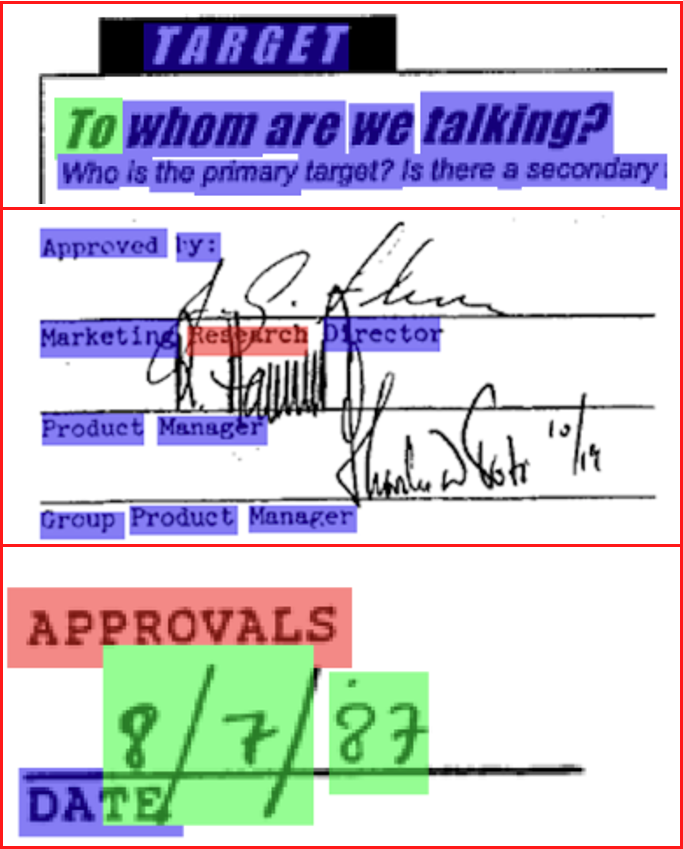}}\quad
	\subcaptionbox{\textbf{\papertitle} multi-modal}[.295\linewidth][c]{%
		\includegraphics[width=\linewidth]{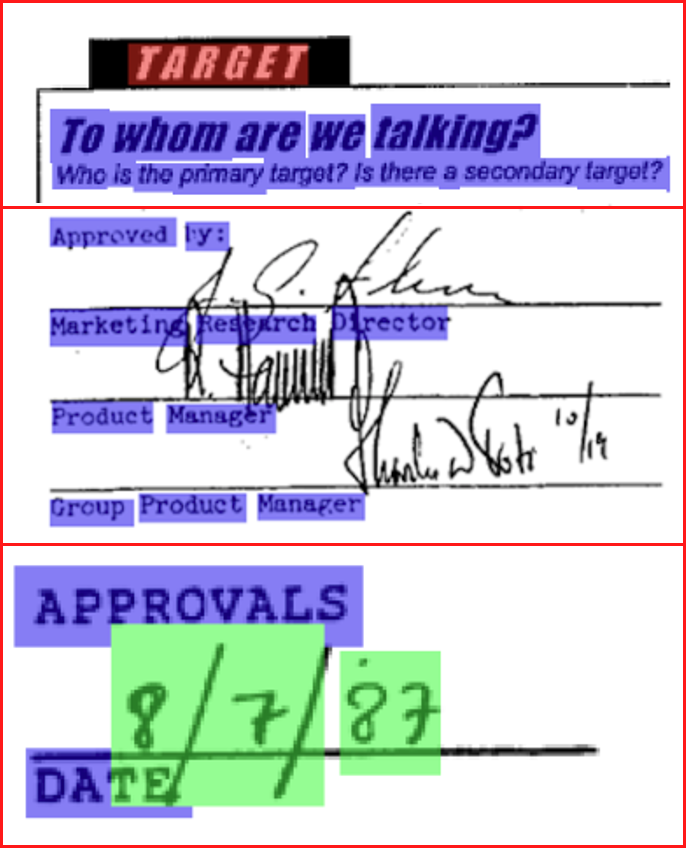}}

    \vspace{-0.8em}
	\caption{ \textbf{\papertitle Qualitative Examples}: From \papertitle on FUNSD test-set \papertitle 83.34 F1 vs LayoutLMv1 78.66 F1. \textbf{Legend}:  \textcolor{red}{\textbf{Red}}: \textbf{Header}-label, \textcolor{blue}{\textbf{Blue}}: \textbf{Question}, \textcolor{green}{\textbf{Green}}: \textbf{Answer}. \textbf{Row 1}: ``TARGET'' is a \textbf{Header}-label which is very visual in nature. \papertitle correctly classifies it whereas a text + spatial model misses such visual cues.
	\textbf{Row 2}: This is a challenging scenario. Notice the word ``Research'' behind the signature. Text + spatial model gets confused and mis-classifies ``Research'' as \textbf{Header}, whereas \papertitle figured out that ``Research'' is part of ``Marketing Research Director'' in spite of visual occlusions. \textbf{Row 3}: Notice ``Approvals'' is partially hidden behind DATE. In spite of that \papertitle correctly labelled ``APPROVALS'' as \textbf{Question}, where as text+spatial model incorrectly labels it as \textbf{Header}. Best viewed in color and digitally. Snippets are from FUNSD file 86079776\_9777, 89856243, and 87125460.}
	\label{fig:qualitative_examples}
\end{figure*}

\textbf{\textit{Does a deeper projection head help?}} So far we used a single linear layer for downstream evaluation as  is common practice \cite{he_cvpr2020_moco,chen_arxiv2020_mocov2,chen_icml2020_simclr,chen_arxiv2020_simclrv2,appalaraju2020towards} to compare against prior art. Recent publications \cite{chen_arxiv2020_simclrv2,appalaraju2020towards} in self-supervision show that a deeper projection head with ReLU activation acts as a one-way filter to enrich the representation space. We adapt this practice and see if a deeper projection head (fc $\shortrightarrow$ ReLU $\shortrightarrow$ LayerNorm $\shortrightarrow$ fc) can improve downstream performance. Table \ref{table:analysis_deeperhead} shows that in the \textit{low-to-medium} data regime adding a more powerful projection head is harmful and could lead to over-fitting. For the \textit{medium-to-large} downstream task data regime, adding a deeper projection head is beneficial.

\begin{table}[H]
	\begin{center}
			\scalebox{0.75}{
		\begin{tabular}{l|c|c|c}
			Dataset & Train samples & Linear head (F1) & Deeper head (F1) \\
			\Xhline{2.5\arrayrulewidth}
			FUNSD \cite{jaume2019} & 149 & 83.34 & 82.93 \textcolor{red}{(-0.41)}\\
			CORD \cite{park2019cord} & 800 & 96.33 & 96.87 \textcolor{forestgreen}{(+0.54)} \\
			RVL-CDIP \cite{harley2015icdar} & 320,000 & 96.17 & 96.85 \textcolor{forestgreen}{(+0.68)} \\
		\end{tabular}
	}
	\end{center}
	\vspace{-3ex}
	\caption{\textbf{Deeper Projection Head}
	}
	\label{table:analysis_deeperhead}
\end{table}



\subsection{Ablation Study}

Since it takes a long time to pre-train on the entire 5M pages and to minimize environmental impact \cite{Henderson2020TowardsTS}, we do all ablation experiments in Table \ref{table:ablation_pretrain} and \ref{table:ablation_architecture} by pre-training with only 1M documents for 5 epochs. In both Table \ref{table:ablation_pretrain} and \ref{table:ablation_architecture},  we show performance in addition to the previous row in the table. Impact due to adding that component is shown in brackets.
We can see in Table \ref{table:ablation_pretrain} that each of our pre-training tasks have something to contribute to the downstream task performance. The contribution also seem to vary depending on the nature of the downstream task.

\begin{table}[H]
	\begin{center}
		\scalebox{0.85}{
			\begin{tabular}{l|c|c}
				Pre-training task & FUNSD (F1) & CORD (F1) \\
				\Xhline{2.5\arrayrulewidth}
				\papertitle + MLM & 72.40 & 90.58 \\
				\papertitle + MM-MLM & 73.91 \textcolor{forestgreen}{(+1.51)} & 90.98  \textcolor{forestgreen}{(+0.4)} \\
					\hspace{3mm} + Learn to Reconstruct (LTR) & 74.68 \textcolor{forestgreen}{(+0.77)} & 92.61  \textcolor{forestgreen}{(+1.63)} \\ 
					\hspace{3mm} + Text describes Image (TDI) & 76.90 \textcolor{forestgreen}{(+2.23)} & 93.36  \textcolor{forestgreen}{(+0.75)} \\
				    \hspace{3mm} final (\papertitle) & 76.90 & 93.36 \\
			\end{tabular}
		}
	\end{center}
	\vspace{-3ex}
	\caption{\textbf{Ablation on pre-training tasks}: We show the impact of various pre-training tasks on two downstream tasks. MLM: masked language modeling \cite{devlin2018bert}. MM-MLM: multi-modal MLM described in Section \ref{section:approach}.}
	\label{table:ablation_pretrain}
\end{table}

\textbf{\papertitle architecture ablation}: In this ablation we look at the impact of various architectural components of \papertitle. Depending on the down-stream task the impact of the proposed multi-modal self-attention varies from 3.89\% to 1.08\%. This shows that the proposed architecture has indeed learned to fuse multiple modalities.

\begin{table}[H]
	\begin{center}
		\scalebox{0.85}{
			\begin{tabular}{l|c|c}
				Model / Component & FUNSD (F1) & CORD (F1) \\
				\Xhline{2.5\arrayrulewidth}
				Text only model (BERT-base) & 61.56 & 89.23 \\
				\hspace{3mm} + spatial features & 73.01 \textcolor{forestgreen}{(+11.45)} & 92.28 \textcolor{forestgreen}{(+3.05)} \\
				\hspace{3mm} + multi-modal self-attention & 76.90 \textcolor{forestgreen}{(+3.89)} & 93.36 \textcolor{forestgreen}{(+1.08)} \\
				\hspace{3mm} final (\papertitle) & 76.90 & 93.36 \\
			\end{tabular}
		}
	\end{center}
	\vspace{-3ex}
	\caption{\textbf{Ablation on \papertitle Components}: We show the impact of various architectural components used in \papertitle on two downstream tasks (FUNSD and CORD).}
	\label{table:ablation_architecture}
\end{table}



{\bf Qualitative Analysis:}
We share some qualitative examples of the predictions from \papertitle. Figure \ref{fig:qualitative_examples} shows some sequence labeling predictions on the FUNSD dataset. (more examples are in the supplemental).

\section{Conclusion}
In this work, we present \papertitle, a multi-modal end-to-end trainable transformer based model for various Visual Document Understanding tasks. We presented the novel multi-modal attention and two novel vision-plus-language pre-training tasks that allows \papertitle to learn effectively without labeled supervision.
We have shown experimentally that \papertitle indeed learns generalized features through its unsupervised pre-training by matching or surpassing existing state-of-the-art results on 4 datasets that cover a variety of document types. We emphasize that \papertitle showed superior performance against strong baselines in-spite of being one of the smallest model (in terms of \# of parameters) in its class. 

In the future, we plan to improve DocFormer's generalizability in multi-lingual settings as well as for more document types such as info-graphics, maps, and web-pages.

\section{Supplemental}


This is the supplemental material for the main DocFormer paper \cite{appalaraju2021docformer}. Please read the main paper for model formulation, performance numbers on various datasets and further analysis and ablation. 

\subsection{Implementation Details} We present all the hyper-parameters in Table \ref{table:implementation_full_1} used for pre-training and fine-tuning \papertitle. We fine-tune on downstream tasks on the same number of epochs as prior art \cite{xu2020layoutlm,xu2020layoutlmv2,bros2020hong}: FUNSD \cite{jaume2019}, Kleister-NDA \cite{gralinski2020kleister} datasets were fine-tuned for 100 epochs. CORD \cite{park2019cord} for 200 epochs. RVL-CDIP \cite{harley2015icdar} for 30 epochs. For Key, Query 1-D relative local attention we choose a span of $8$ i.e. for a particular multi-modal feature, \papertitle gives more attention $8$ tokens to its left and right.

\begin{table}[H]
	\begin{center}
	\scalebox{0.85}{
		\begin{tabular}{l|c|c}
			Hyper-Parameter & Pre-training & Fine-tuning \\
			\Xhline{2.5\arrayrulewidth}
			Epochs & 5 & varies \\
			Learning rate & 5E-05 & 2.5E-05 \\
			Warm-up & 10\% iters & 0 \\
			Gradient Clipping & 1.0 & 1.0 \\
			Gradient agg. & False & False \\
			Optimizer & AdamW\cite{loshchilov2017decoupled} & AdamW\cite{loshchilov2017decoupled} \\
			Lower case & True & True  \\
			Sequence length & 512 & 512 \\
			Encoder layers & 12 & 12 \\
			32-bit mixed precision & True & True \\
			Batch size & 9 per GPU & 4 per GPU \\
			GPU hardware & A100 (40GB) & V100 (16GB) \\
			Training Num. Samples & 5M & varies \\
			Training time & 17 hours/epoch & varies \\
		\end{tabular}
		}
	\end{center}
	\vspace{-3ex}
	\caption{\textbf{Implementation Details}: Hyper-parameters used for pre-training \papertitle and fine-tuning for downstream tasks. Training epochs vary for down-stream tasks.}
	\label{table:implementation_full_1}
	\vspace{-1ex}
\end{table}

\subsection{Run-time Complexity Analysis}
Since we propose a variant of the self-attention \cite{vaswani2017attention} operation, we compute the train and inference run-time analysis in big-o notation.

\begin{table}[H]
	\begin{center}
	\scalebox{0.85}{
		\begin{tabular}{l|c|c}
			Layer Type & Run-time Complexity & Seq. Complexity \\
			\Xhline{2.5\arrayrulewidth}
			Convolution & $O\left(k \cdot n \cdot d^{2}\right)$ & $O(1)$  \\
			Recurrent & $O\left(n \cdot d^{2}\right)$ & $O(n)$ \\
			Self-Attention & $O\left(n^{2} \cdot d\right)$ & $O(1)$ \\
			Self-Attention (relative) & $O\left(r \cdot n \cdot d\right)$ & $O(1)$ \\
			\papertitle MMSA & $2 \cdot [ O\left(n^{2} \cdot d\right) ]$ & $O(1)$ \\
		\end{tabular}
		}
	\end{center}
	\vspace{-3ex}
	\caption{\textbf{Complexity analysis}: Here $n$ is the sequence length, $d$ is the representation dimension, $k$ is the kernel
size of convolutions and $r$ the size of the neighborhood in restricted self-attention. Omitting number of attention heads $h$ for brevity. Here assume $h=1$. In addition, MMSA: multi-modal self-attention.}
	\label{table:implementation_full}
	\vspace{-1ex}
\end{table}

Please note that the full run-time complexity for \papertitle has been abridged as the self-attention is the most significant operation (keeping in line with big-O notation). In addition, the presence of $2$ is to signify the unique MMSA operation proposed in this paper, where multi-modal feature from each layer is added with image and spatial features (see Section 3.1). We see that DocFormer's multi-modal self-attention (Section 3.1) is an efficient way to do multi-modal learning.

\subsection{Pseudo-code}
We present a rough pseudo-code for our novel multi-modal self-attention (MMSA) as described in section 3.1. We believe the pseudo-code would aid an independent researcher to better replicate our proposed novelty. Please note omitting dropout and layer norm at the end for brevity.

\begin{lstlisting}[language=Python]
##Multi Modal Self Attention

#text kqv embeddings
key1 = Linear(d_model, n_head * d_k)
query1 = Linear(d_model, n_head * d_k)
value1 = Linear(d_model, n_head * d_v)

#image kqv embeddings
key2 = Linear(d_model, n_head * d_k)
query2 = Linear(d_model, n_head * d_k)
value2 = Linear(d_model, n_head * d_v)

#spatial embeddings. note! shared by text, image
key3 = Linear(d_model, n_head * d_k)
query3 = Linear(d_model, n_head * d_k)

#See Eq. 6 and 7 in main paper for formulation
def multi_modal_self_attention(emb, img_feat, spatial_feat):

  #self-attention of text (and prev. layers subseq.)
  k1,q1,v1 = emb,emb,emb
  k1 = rearr(key1(k1), 'b t (head k) -> head b t k')
  q1 = rearr(query1(q1), 'b l (head k) -> head b l k')
  v1 = rearr(value1(v1), 'b t (head v) -> head b t v')
  attn1 = einsum('hblk,hbtk->hblt', [q1,k1])/sqrt(q1.shape[-1])

  #1D relative pos. (query, key)
  #note rel_pos_embed1 is learnt relative pos emb. nxn
  rel_pos_key1 = einsum('bhrd,lrd->bhlr', k1, rel_pos_embed1)
  rel_pos_query1 = einsum('bhld,lrd->bhlr', q1, rel_pos_embed1)

  #shared spatial - text/hidden features
  sp_k1, sp_q1 = spatial_feat, spatial_feat
  sp_k1=rearr(key3(sp_k1),'b t (head k) -> head b t k')
  sp_q1=rearr(query3(sp_q1),'b l (head k)->head b l k')
  text_only_spatial_scores = einsum('hblk,hbtk->hblt', [sp_q1,sp_k1])/sqrt(sp_q1.shape[-1])

  text_attn_scores = attn1 + rel_pos_key1 + rel_pos_query1 + text_only_spatial_scores

  #-----
  ##Self-attn of image (repeat of above for img feat)
  k2,q2,v2 = img_feat,img_feat,img_feat
  k2 = rearr(key2(k2), 'b t (head k) -> head b t k')
  q2 = rearr(query2(q2), 'b l (head k) -> head b l k')
  v2 = rearr(value2(v2), 'b t (head v) -> head b t v')
  attn2 = einsum('hblk,hbtk->hblt', [q2,k2])/sqrt(q2.shape[-1])

  #1D relative pos. (query, key)
  #note rel_pos_embed1 is learnt relative pos emb. nxn
  rel_pos_key2 = einsum('bhrd,lrd->bhlr', k2, rel_pos_embed2)
  rel_pos_query2 = einsum('bhld,lrd->bhlr', q2, rel_pos_embed2)

  #shared spatial - img features
  sp_k2, sp_q2 = spatial_feat, spatial_feat
  sp_k2=rearr(key3(sp_k2),'b t (head k) -> head b t k')
  sp_q2=rearr(query3(sp_q2),'b l (head k)->head b l k')
  img_only_spatial_scores = einsum('hblk,hbtk->hblt', [sp_q2,sp_k2])/sqrt(sp_q2.shape[-1])

  img_attn_scores = attn2 + rel_pos_key2 + rel_pos_query2 + img_only_spatial_scores

  #---- attended output: multi-modal
  text_attn_probs = dropout(softmax(dim=-1)(text_attn_scores))
  img_attn_probs = dropout(softmax(dim=-1)(img_attn_scores))
  
  text_cntx = einsum('hblt,hbtv->hblv', [text_attn_probs, v1])
  img_cntx = einsum('hblt,hbtv->hblv', [img_attn_probs, v2])
  context = text_cntx + img_cntx
  return context
\end{lstlisting}

\subsection{\papertitle Architecture for Downstream Tasks}
\begin{figure*}[hbt!]
	\centering
	\subcaptionbox{Architecture for downstream Document Classification Task}[.45\linewidth][c]{%
	\includegraphics[width=\linewidth]{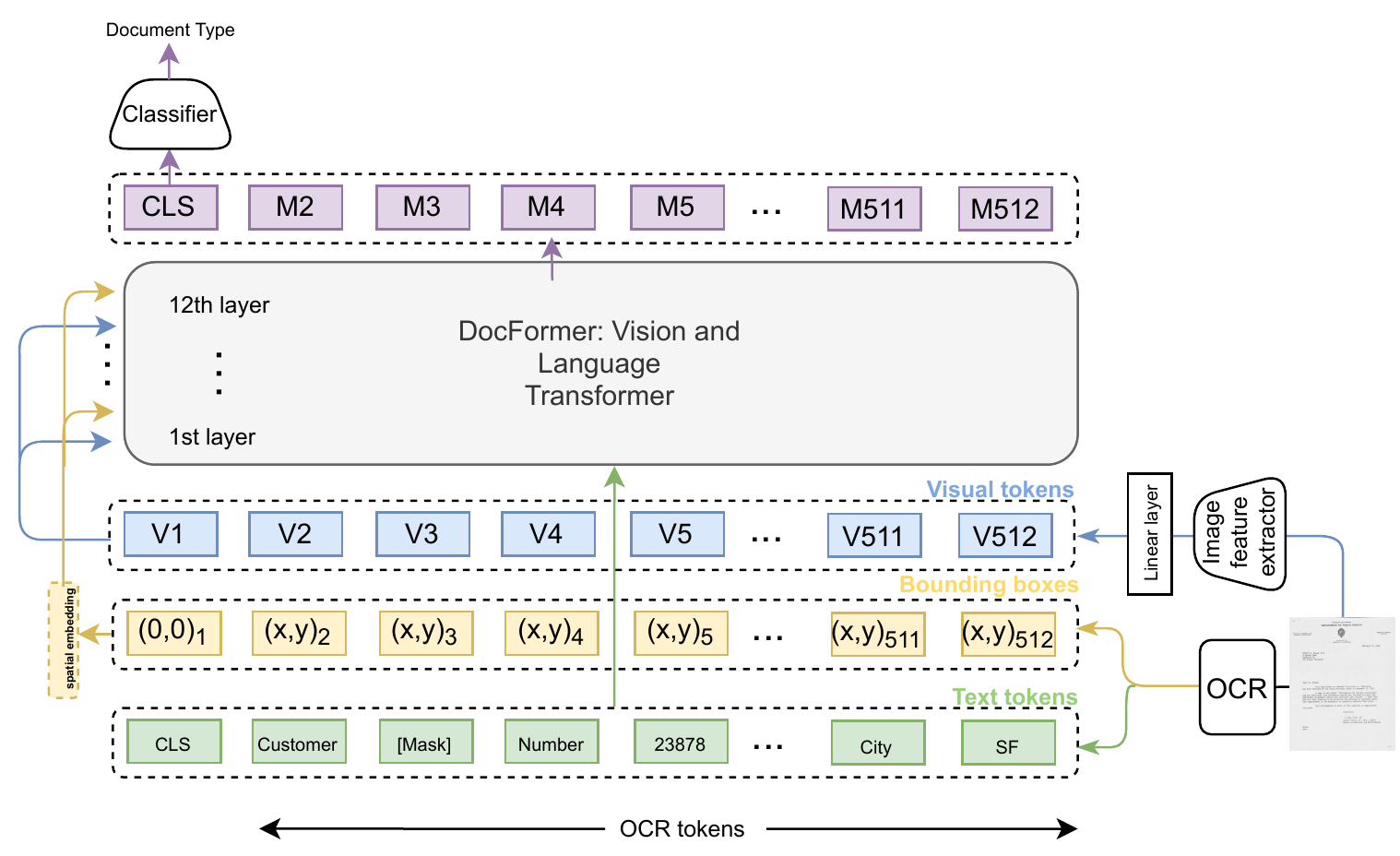}}\quad
	\subcaptionbox{Architecture for downstream Sequence and Entity Labeling Tasks}[.45\linewidth][c]{%
	\includegraphics[width=\linewidth]{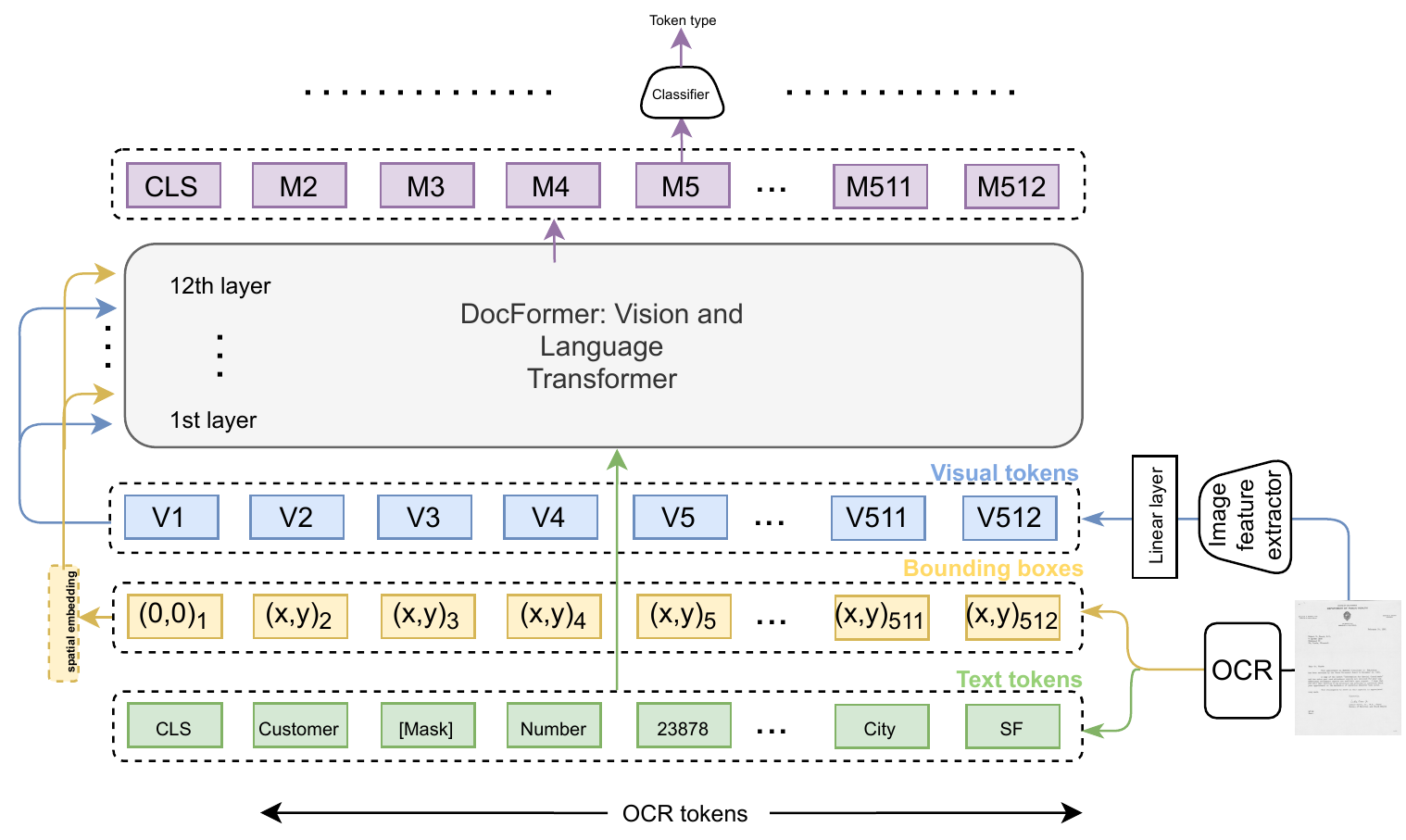}}\quad
	\caption{\textbf{\papertitle architecture for various downstream tasks}: Image on \textbf{Left} (a) is the architecture for document classification [CLS] is a pooling layer (fn $\shortrightarrow$ ReLU $\shortrightarrow$ fn) to get a pooled representation used for document classification task. Image on \textbf{Right} (b) is the architecture used for entity and sequence labeling tasks. Note, only a single linear layer is added for all downstream tasks. Also, all components of \papertitle are fine-tuned for each of the downstream tasks.}
	\label{fig:downstream_arch}
\end{figure*}

\papertitle is pre-trained as mentioned in section 3.2. After training it for 5 epochs, we remove the pre-training multi-task heads and use \papertitle (including the visual branch) as a backbone. We simply add a trainable linear-head which predicts the appropriate number of classes which is dataset specific. 
Please see Figure \ref{fig:downstream_arch} for architecture modifications for downstream tasks.

\subsection{\papertitle Multi-Modal Self-Attention} 
\begin{figure*}[hbt!]
\begin{center}
\includegraphics[width=1.0\linewidth]{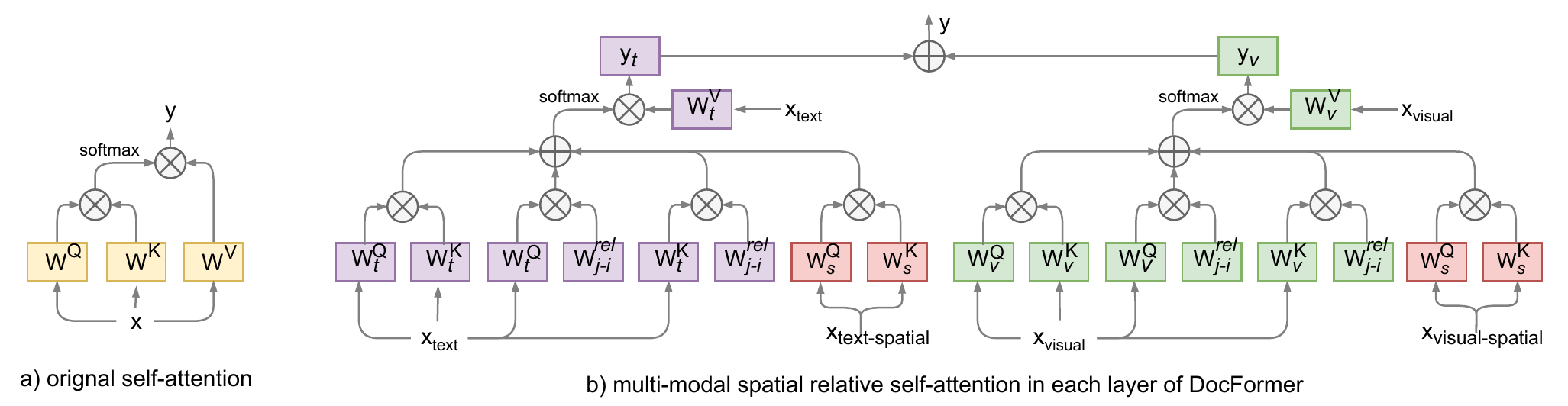}
\end{center}
   \caption{\textbf{Multi-Modal Self-Attention Layer}: the image \textbf{a) Left} shows the traditional self-attention proposed in Vaswani et al \cite{vaswani2017attention}. Note the multi-head attention and feed-forward layers are omitted for brevity. Cross (X) is matrix-multiplication and (+) is element-wise addition. \textbf{b) Right} shows the proposed multi-modal self-attention layer. This comprises each layer of \papertitle. Notice, the spatial weights across text, vision are shared (RED color), thus helping \papertitle address the \textit{cross-modality feature correlation} issue commonly faced in multi-modal training. The notation is consistent with Equations 1-7 in the main paper. Best viewed in color.}
\label{fig:mm_self_attn_supp}
\end{figure*}

In Figure \ref{fig:mm_self_attn_supp} we show a more detailed visual representation of the novel multi-modal self-attention introduced in this paper.  For reference we also show the original self-attention used by Vaswani et al. \cite{vaswani2017attention}.

\subsection{FUNSD Vizualizations}
\papertitle achieves state-of-the-art performance of 83.34\% F1-score (see Section 4.1) on FUNSD \cite{jaume2019} dataset amongst other multi-modal models its size. In this sub-section we look at more visualizations by \papertitle on the test-set. One important aspect of this VDU we would like to mention is the OCR is not in human reading-order.

\begin{figure*}
	\centering
    \subcaptionbox{Ground Truth}[.47\linewidth][c]{%
	\includegraphics[width=\linewidth]{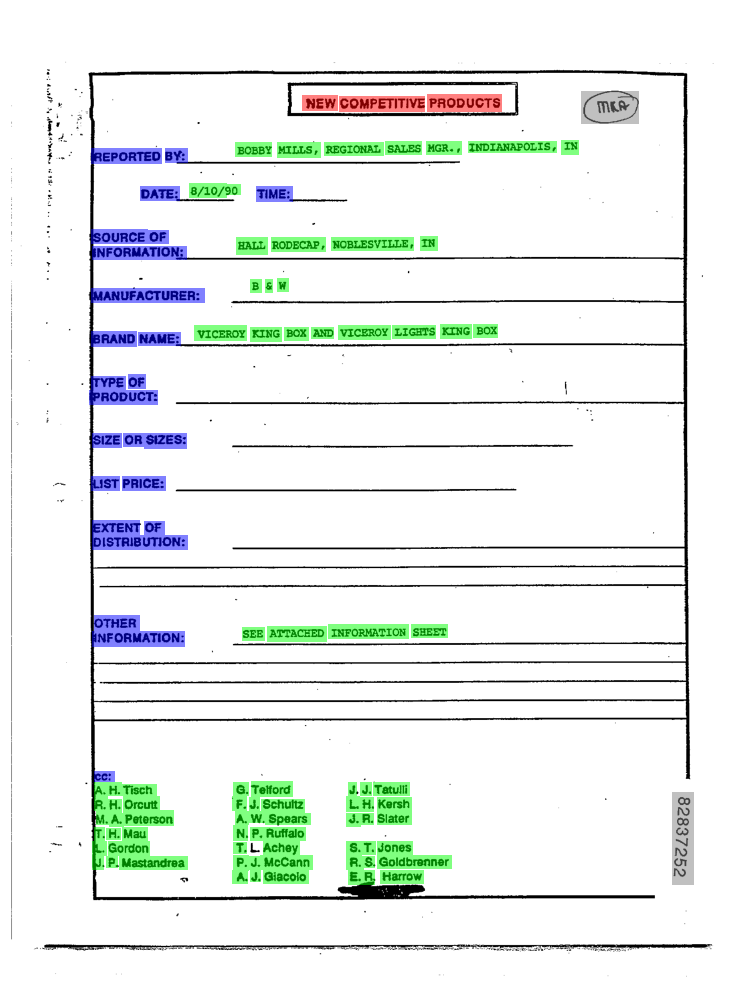}}\quad
	\subcaptionbox{\papertitle predictions}[.47\linewidth][c]{%
	\includegraphics[width=\linewidth]{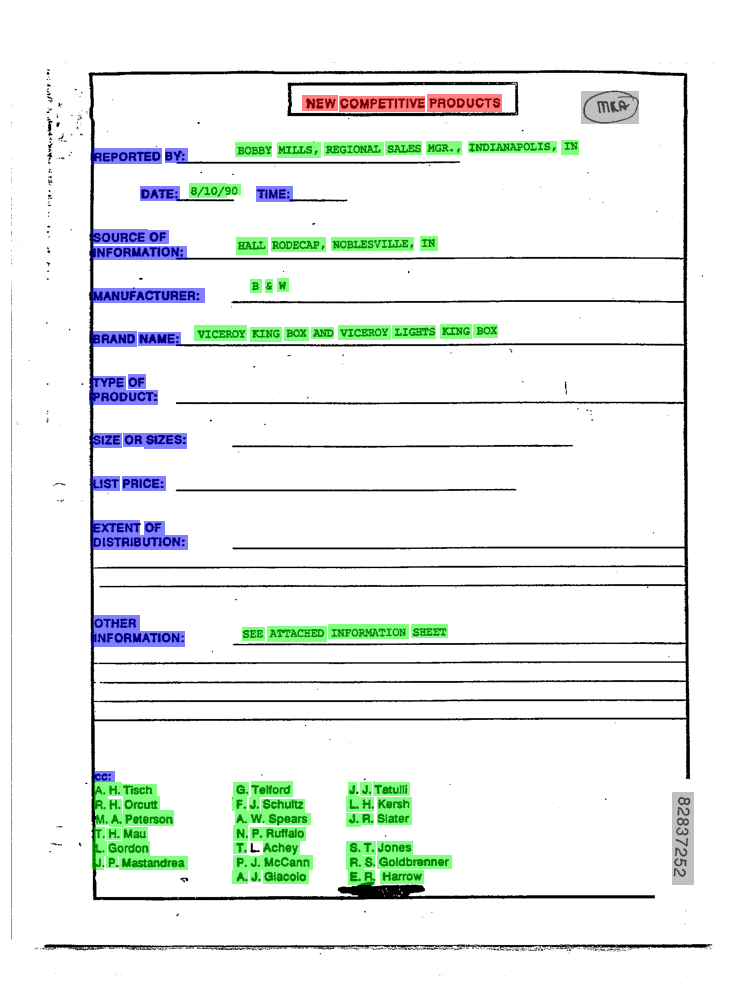}}\quad

	\caption{\textbf{\papertitle perfect predictions for 82837252 testfile of FUNSD dataset}: Left image shows GT and right image is the prediction made by \papertitle which perfectly matches with GT. Best viewed in color.} 

	\label{fig:funsd_supplemental_viz3}
\end{figure*}

\begin{figure*}[hbt!]
	\centering
	\subcaptionbox{Ground Truth}[.47\linewidth][c]{%
	\includegraphics[width=\linewidth]{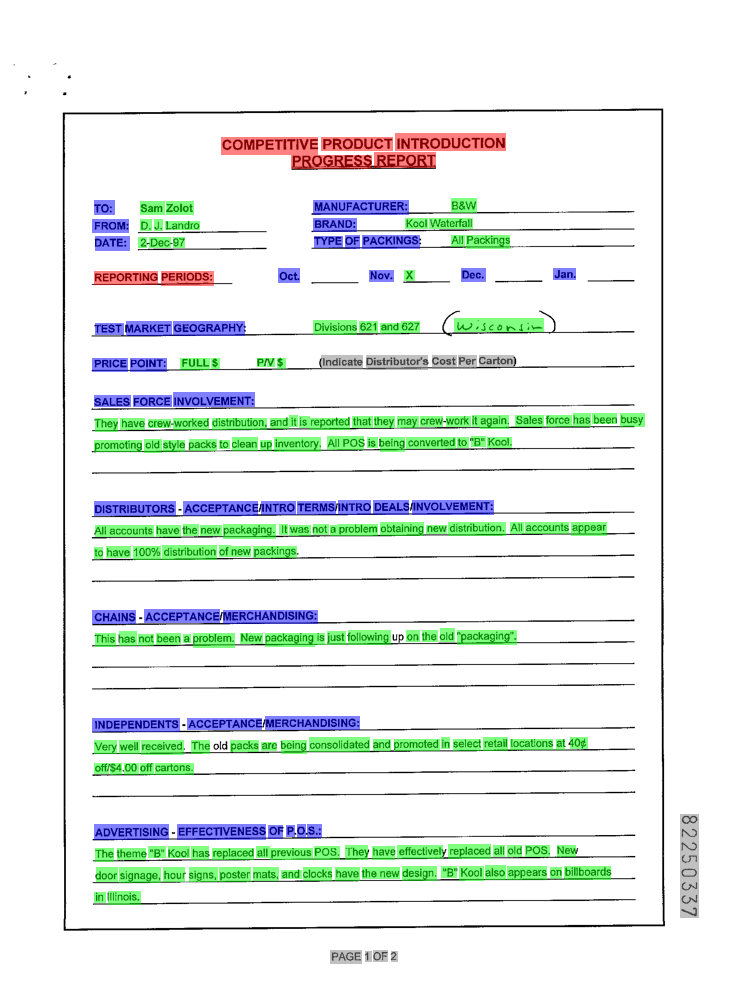}}\quad
	\subcaptionbox{\papertitle predictions}[.47\linewidth][c]{%
	\includegraphics[width=\linewidth]{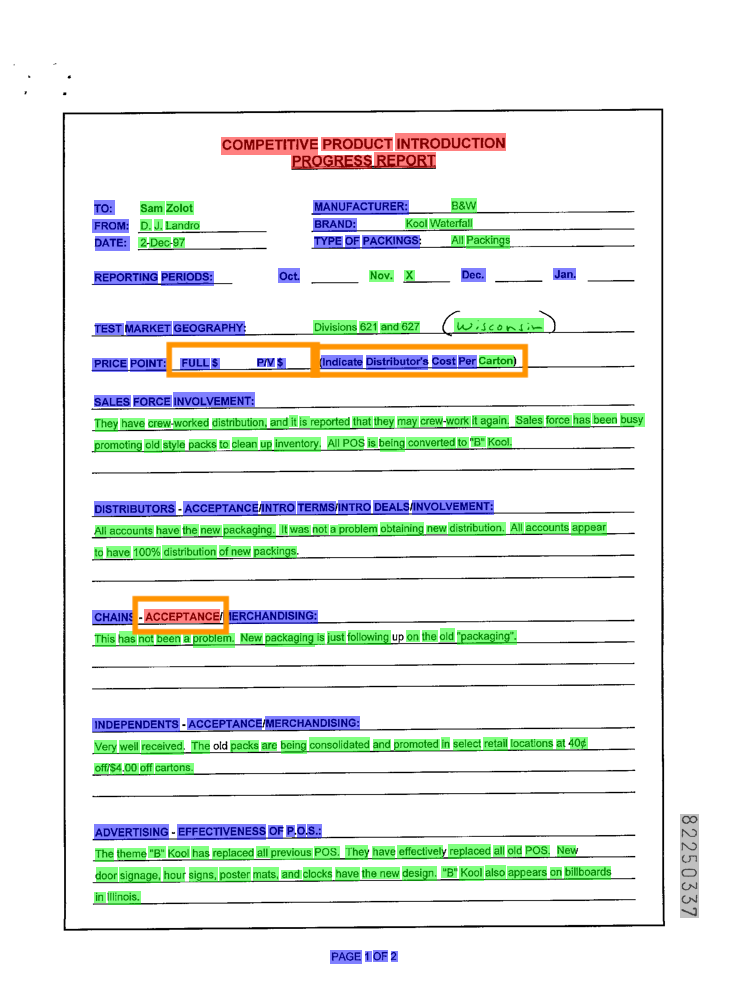}}\quad
	\caption{\textbf{\papertitle slightly bad predictions for 82250337\_0338 test\-file on FUNSD dataset}: Based on the predictions on the right (b), we can see that \papertitle was able to classify most of the sequence correctly. However, if we look at the orange bounding boxes we can spot the errors. "(Indicate Distributor's Cost per Carton)" is tagged as Other text in ground-truth but \papertitle incorrectly classified part of the tokens as \textcolor{blue}{\textbf{Question}}. Best if viewed digitally and in color.}
	\label{fig:funsd_supplemental_viz1}
\end{figure*}

\begin{figure*}
	\centering
	\subcaptionbox{Ground Truth}[.47\linewidth][c]{%
	\includegraphics[width=\linewidth]{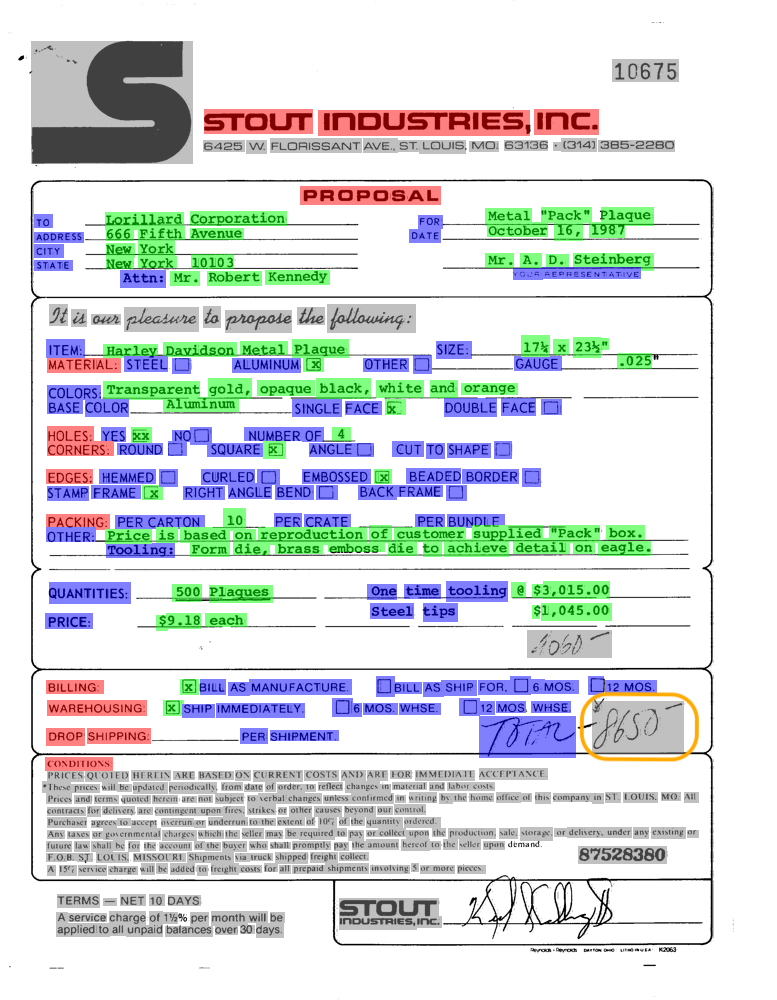}}\quad
	\subcaptionbox{\papertitle predictions}[.47\linewidth][c]{%
	\includegraphics[width=\linewidth]{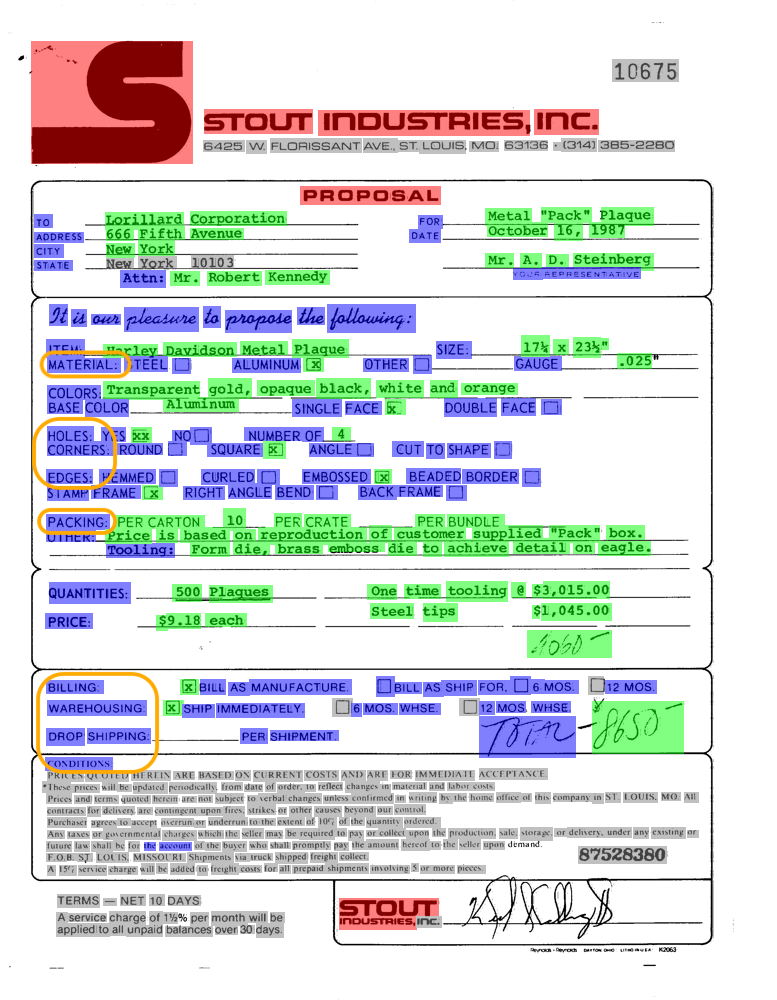}}\quad
	\caption{\textbf{\papertitle slightly bad predictions for 87528380 testfile on FUNSD dataset}: Here, we focus the readers attention on two specific scenarios: FUNSD dataset has been known to have ground-truth annotation issues. We  find on the left image the orange highlighted box $"8650"$ is incorrectly annotated in GT as $"other"$ text, however \papertitle correctly predicts it as "answer" token for the question $"total"$. \textbf{Scenario 2:} The orange highlighted boxes on the right image are tokens which are actually sub-headers but \papertitle mis-classifies as $"question"$ tokens. In this case, \papertitle likely gave more weight-age to language features and not so much to visual features and so ended up mis-classifying. We would like to point out that this is an ambiguous example as the language in mis-classified regions do look like $"questions"$. Best viewed in color. }
	\label{fig:funsd_supplemental_viz2}
\end{figure*}

Please note that, we search for and present cases where mistakes were made by \papertitle with the aim of understanding mistakes. Legend for the colors used in images is, Header-label: \textcolor{red}{\textbf{Red}}, Question: \textcolor{blue}{\textbf{Blue}}, Answer: \textcolor{green}{\textbf{Green}}, Other: Grey color. Please see Figures \ref{fig:funsd_supplemental_viz3}, \ref{fig:funsd_supplemental_viz1}, \ref{fig:funsd_supplemental_viz2}.

\begin{figure*}
	\centering
    \subcaptionbox{Example FUNSD document with Ground Truth overlays}[.40\linewidth][c]{%
	\includegraphics[width=\linewidth]{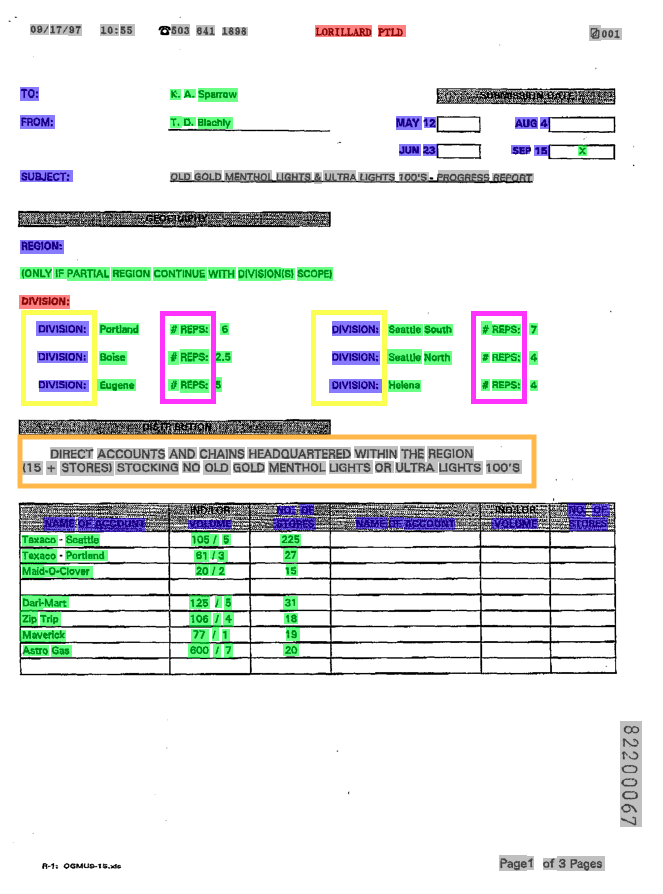}}\quad
	\subcaptionbox{\papertitle prediction self-attention heatmap (last encoder layer, 2nd head). DocFormer has up to 512 tokens in each layer. Each point on the image shows the strength of attention from a token on the $y$-axis to a token on the $x$-axis. The bluer colors show more attention and the red less attention.
	}[.40\linewidth][c]{%
	\includegraphics[width=\linewidth]{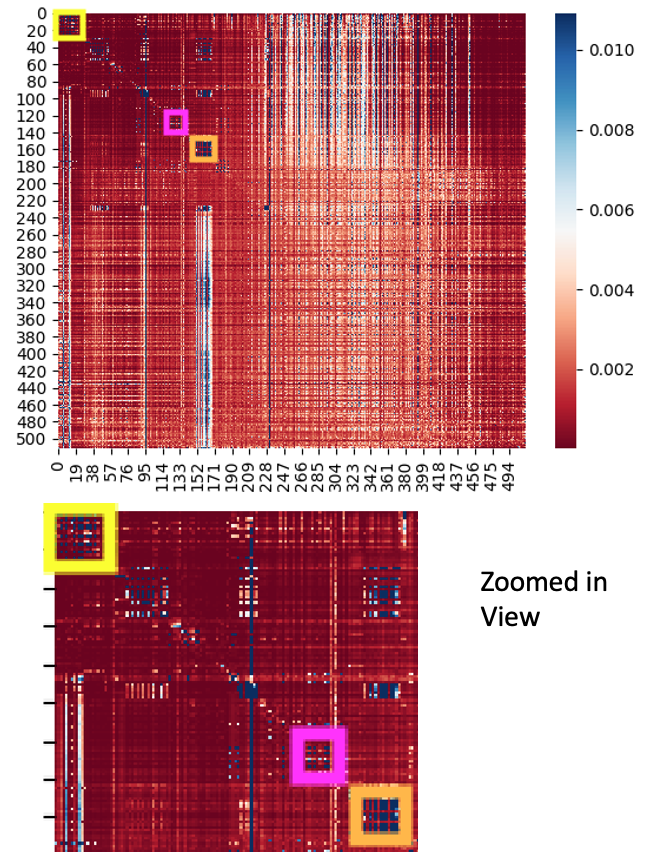}}\quad

	\caption{\textbf{\papertitle learns repetition and regularity}: the \textcolor{yellow}{\textbf{yellow}} and \textcolor{magenta}{\textbf{purple}} boxes in the left figure matches the  \textcolor{yellow}{\textbf{yellow}} and \textcolor{magenta}{\textbf{purple}} boxes in the right figure. The OCR is not in reading order. Hence the six occurrences of ``DIVISION" appear together in front (among the top 25) - \textcolor{yellow}{\textbf{yellow}} box in Figure b) and they correspond to the \textcolor{yellow}{\textbf{yellow}} boxes in Figure a).
	Similarly, the \textcolor{magenta}{\textbf{purple}} box in Figure b) corresponds to the \textcolor{magenta}{\textbf{purple}} boxes in Figure a).
	\papertitle is able to pick up such repetitions as strong self-attention signals (blue colored pixels in the right self-attention figure) that help the model solve the task. 
	This example shows that regular indentation and spacing help \papertitle understand the form better just as they would help humans parse a form.
	The \textcolor{orange}{\textbf{orange}} boxed region in the heatmap also shows strong self-attention. We think that is due to \papertitle representing the blob of text as a single paragraph (in this case, as background text). Best viewed digitally and in color. }

	\label{fig:funsd_supplemental_atten_map}
\end{figure*}

In Figure \ref{fig:funsd_supplemental_atten_map}, we show one specific pattern that \papertitle learns through its novel multi-modal self-attention. We show that \papertitle automatically learns repetitive local patterns even though it was not explicitly taught this.

\subsection{CORD Visualizations}
\papertitle matches the state-of-the-art performance of 96.33\% F1-score on CORD \cite{park2019cord} dataset (previous state-of-the-art model TILT-large consists of 780M parameters almost 4x the size of \papertitle). Please see Section 4.3 in the main paper.

\begin{figure*}
	\centering
	\subcaptionbox{Ground Truth (left) and \papertitle predictions (right)}[.63\linewidth][c]{%
	\includegraphics[width=\linewidth]{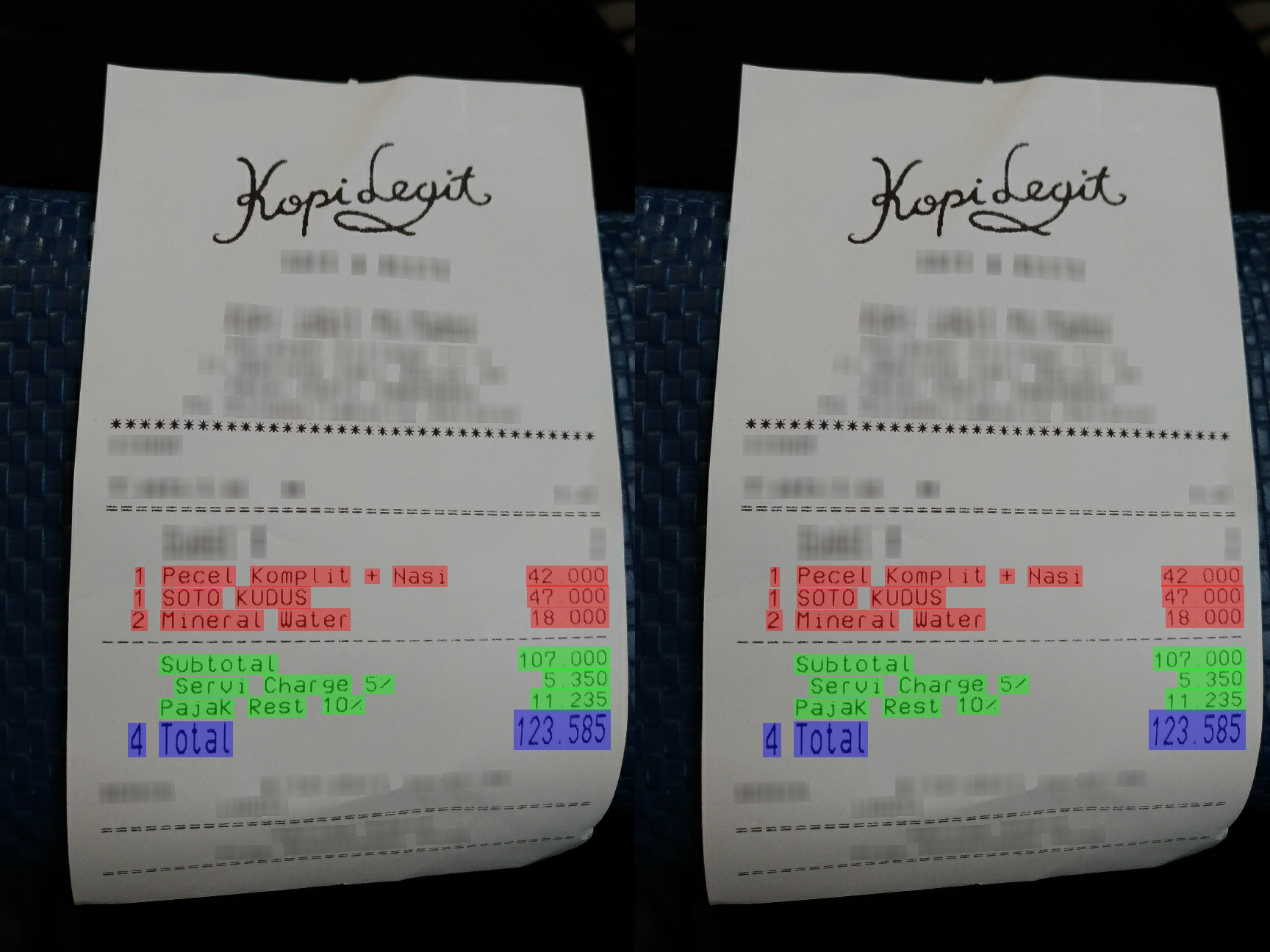}}\quad
	\caption{\textbf{\papertitle predictions on CORD}: For file receipt\_00053 (a) shows both ground-truth and predictions. \papertitle predicted correctly all the entity regions in the image. Best if viewed digitally and in color.}
	\label{fig:cord_supplemental_viz1}
\end{figure*}

\begin{figure*}
	\centering
	\subcaptionbox{Ground Truth (left) and \papertitle predictions (right)}[.63\linewidth][c]{%
	\includegraphics[width=\linewidth]{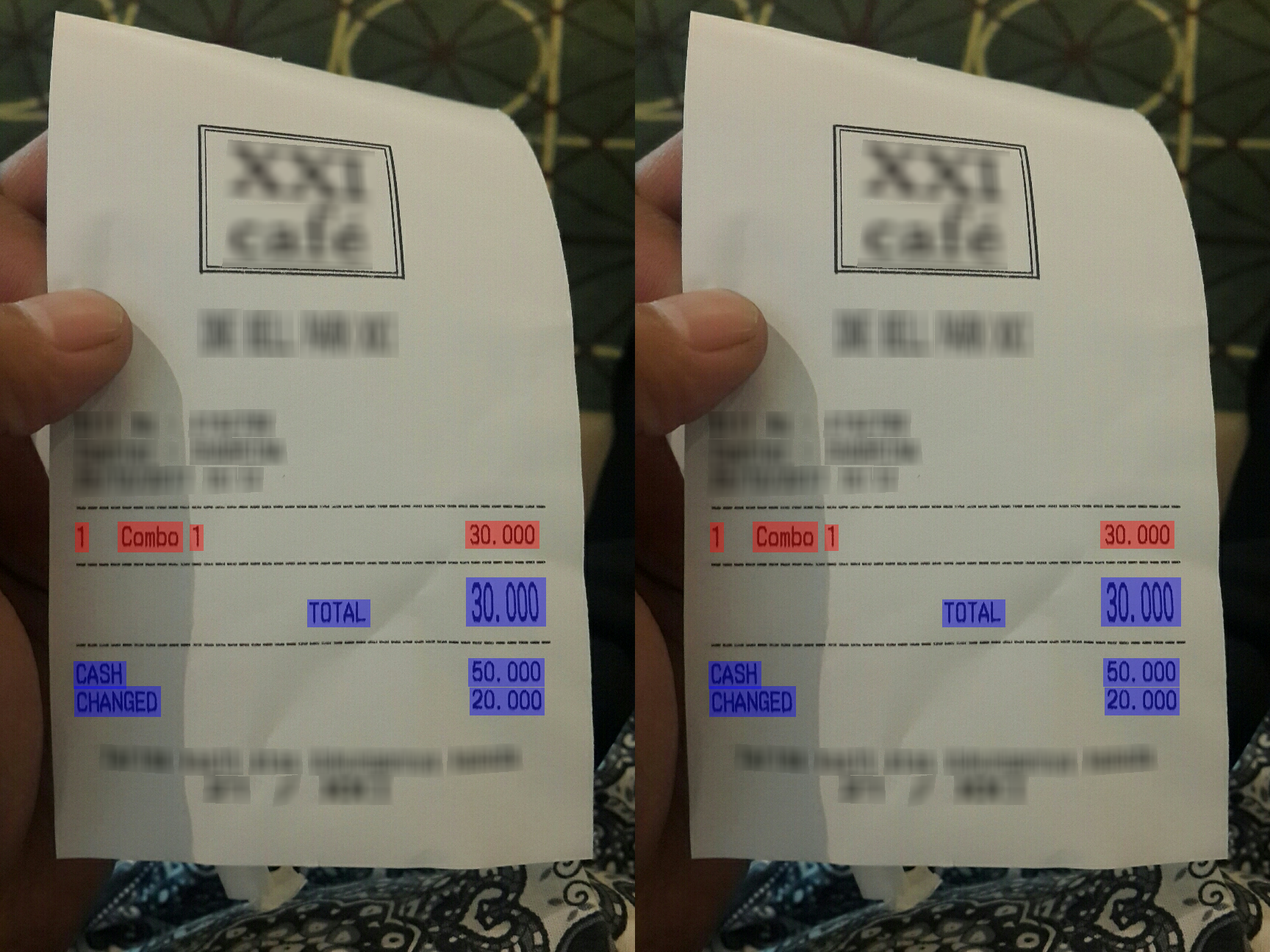}}\quad
	\caption{\textbf{\papertitle predictions on CORD}: For file receipt\_00044. Best if viewed digitally and in color.}
	\label{fig:cord_supplemental_viz1_2}
\end{figure*}

\begin{figure*}
	\centering
	\subcaptionbox{Ground Truth (left) and \papertitle predictions (right)}[.63\linewidth][c]{%
	\includegraphics[width=\linewidth]{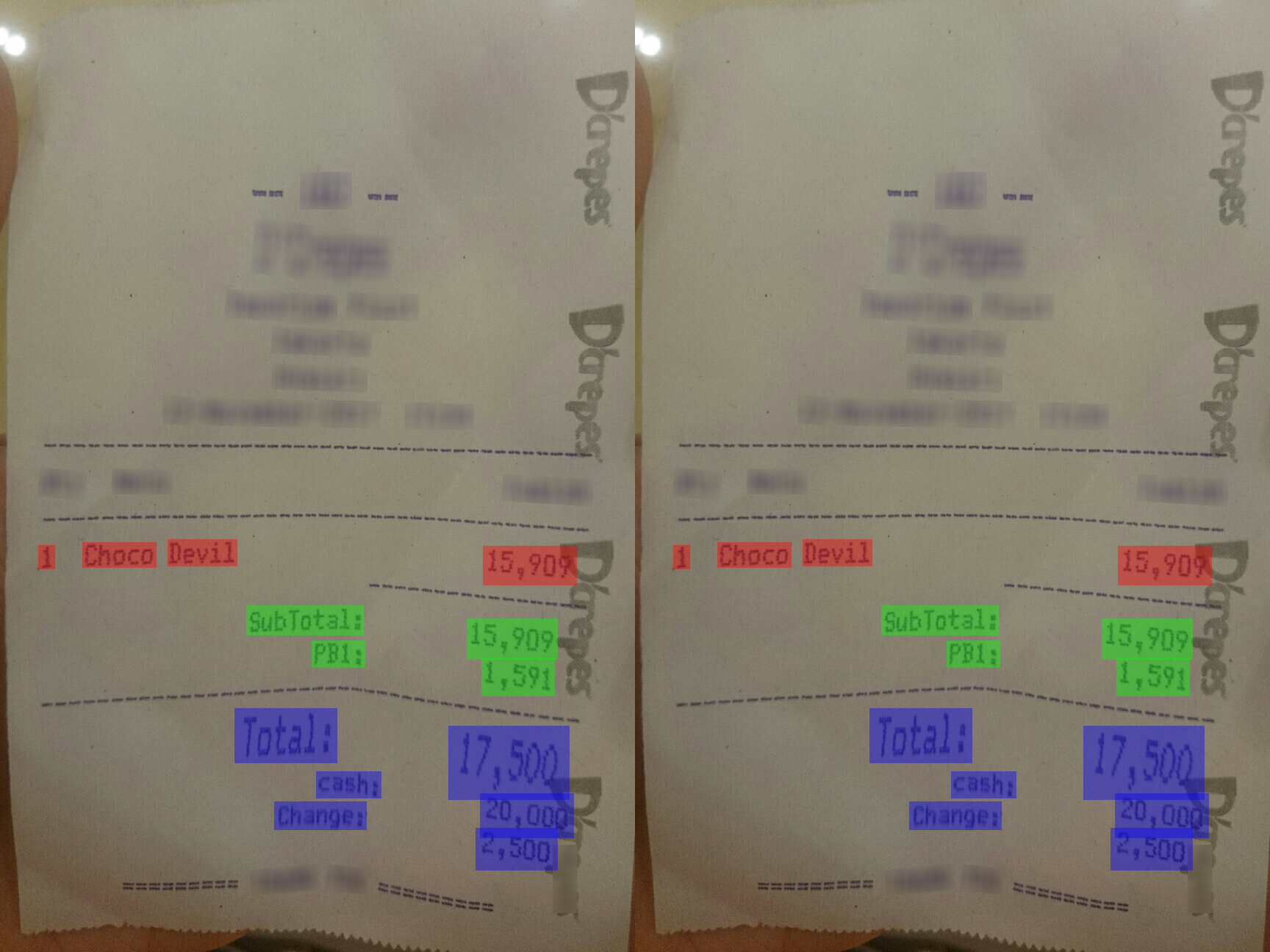}}\quad
	\caption{\textbf{\papertitle predictions on CORD}: For file receipt\_00072. Best if viewed digitally and in color.}
	\label{fig:cord_supplemental_viz1_3}
\end{figure*}

\begin{figure*}
	\centering
    \subcaptionbox{Ground Truth}[.40\linewidth][c]{%
	\includegraphics[width=\linewidth]{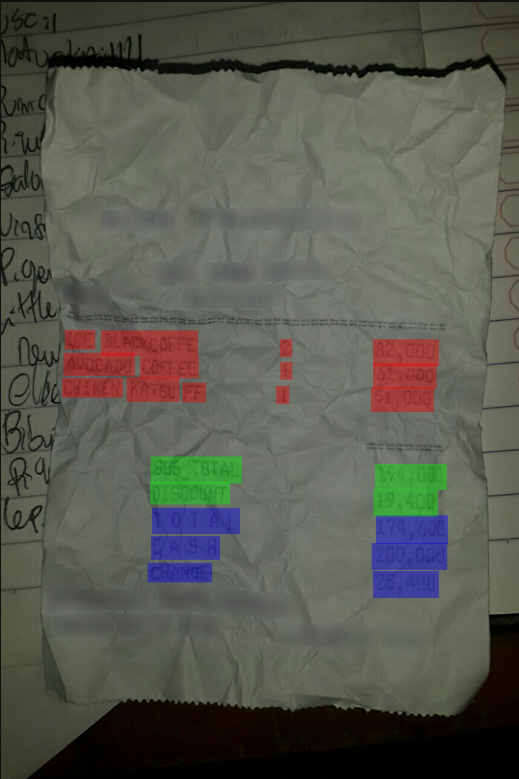}}\quad
	\subcaptionbox{\papertitle predictions}[.40\linewidth][c]{%
	\includegraphics[width=\linewidth]{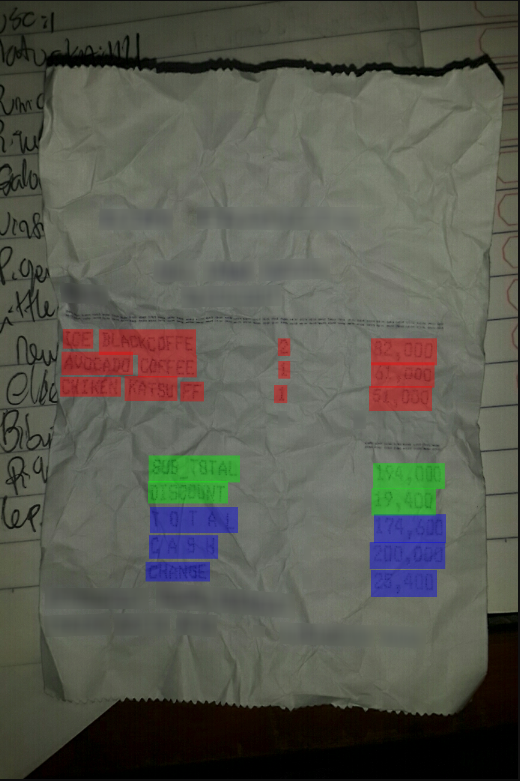}}\quad

	\caption{\textbf{\papertitle perfect predictions on CORD dataset}: Left image shows GT and right image is the prediction for file receipt\_00004 made by \papertitle which perfectly matches with the GT despite the presence of distortion and background text.} 

	\label{fig:cord_supplemental_viz2_1}
\end{figure*}

\begin{figure*}
	\centering
    \subcaptionbox{Ground Truth}[.40\linewidth][c]{%
	\includegraphics[width=\linewidth]{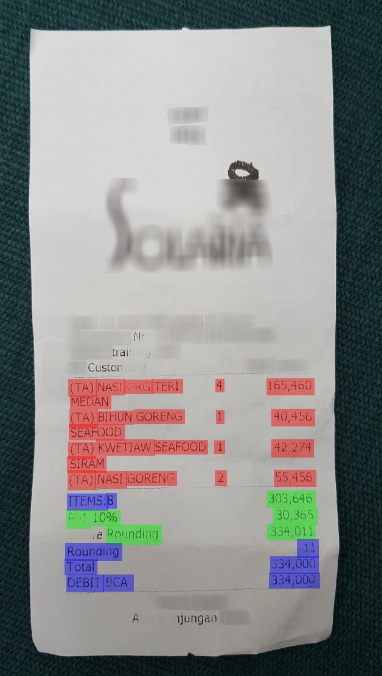}}\quad
	\subcaptionbox{\papertitle predictions}[.40\linewidth][c]{%
	\includegraphics[width=\linewidth]{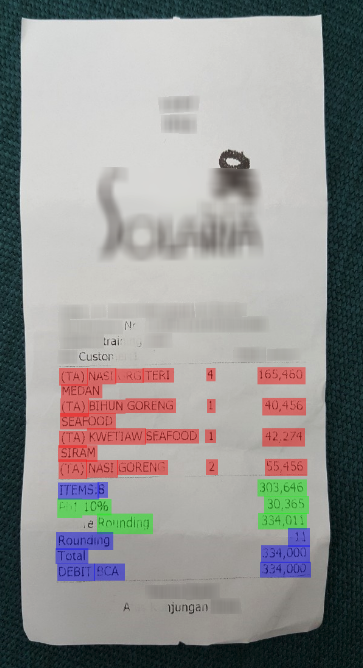}}\quad

	\caption{\textbf{\papertitle perfect predictions on CORD dataset}: Left image shows GT and right image is the prediction for file receipt\_00051 made by \papertitle which perfectly matches with GT. Note that the faded out text which is hard to OCR is correctly classified due to multi-modal self-attention features.} 

	\label{fig:cord_supplemental_viz2_2}
\end{figure*}

\begin{figure*}
	\centering
    \subcaptionbox{Ground Truth}[.40\linewidth][c]{%
	\includegraphics[width=\linewidth]{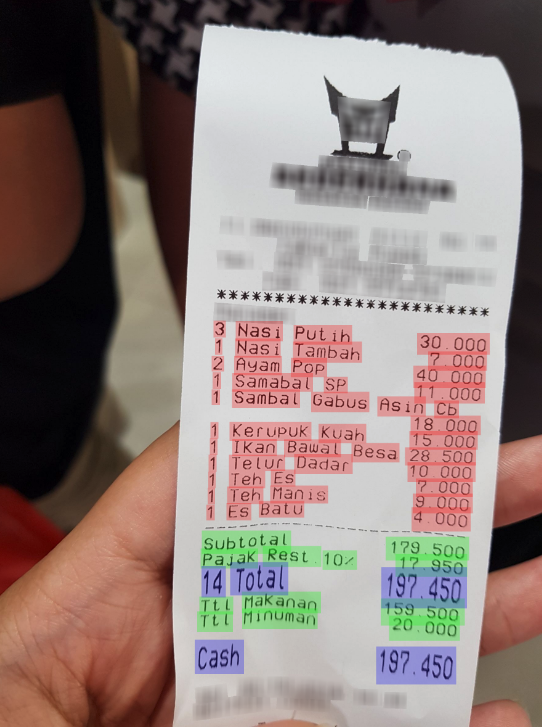}}\quad
	\subcaptionbox{\papertitle predictions}[.40\linewidth][c]{%
	\includegraphics[width=\linewidth]{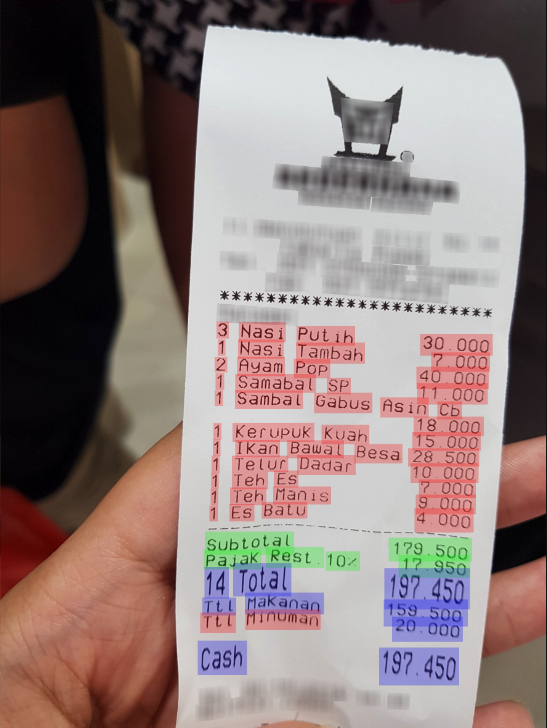}}\quad

	\caption{\textbf{\papertitle Partially correct predictions on CORD dataset}: Left image shows GT and right image is the prediction for file receipt\_00085 made by \papertitle with a misclassification of tokens of category SUBTOTAL with TOTAL items. This could be due to the rarity of SUBTOTAL tokens appearing below TOTAL tokens which \papertitle may not have encountered during training. } 

	\label{fig:cord_supplemental_viz2_3}
\end{figure*}

In this sub-section we look at CORD \cite{park2019cord}  visualizations by \papertitle. We explicitly show hard-cases where \papertitle does well, see Figures \ref{fig:cord_supplemental_viz1}, \ref{fig:cord_supplemental_viz1_2},  \ref{fig:cord_supplemental_viz1_3}, \ref{fig:cord_supplemental_viz2_1}, \ref{fig:cord_supplemental_viz2_2}. In order to be transparent, we also show an error scenario in Figure \ref{fig:cord_supplemental_viz2_3}. Legend for the colors in images is, Menu items: \textcolor{red}{\textbf{Red}}, Total: \textcolor{blue}{\textbf{Blue}}, Sub-total (pre-tax): \textcolor{green}{\textbf{Green}}, Void-menu: Cyan color, Other: grey. 


\clearpage
{\small
\bibliographystyle{ieee_fullname}
\bibliography{egbib}
}

\end{document}